\documentclass[twoside,11pt]{article}

\usepackage{jair}
\usepackage{theapa}

\usepackage{lingmacros}
\usepackage{psfig}
\usepackage{graphicx}
\usepackage[latin1]{inputenc}

\jairheading{30}{2007}{413-456}{05/2007}{11/2007}
\ShortHeadings{Individual and Domain Adaptation in Dialogue}
{Walker, Stent, Mairesse, \& Prasad}
\firstpageno{413}

\title{Individual and Domain Adaptation \\ in Sentence Planning for Dialogue}

 \author{\name Marilyn Walker \email lynwalker@gmail.com \\
        \addr Department of Computer Science, University of Sheffield\\
        211 Portobello Street, Sheffield S1 4DP, United Kingdom
        \AND
        \name Amanda Stent \email amanda.stent@gmail.com\\
        \addr Department of Computer Science, Stony Brook University\\
       Stony Brook, NY 11794, USA
        \AND
        \name Fran\c{c}ois Mairesse \email f.mairesse@sheffield.ac.uk \\
        \addr Department of Computer Science, University of Sheffield,\\
        211 Portobello Street, Sheffield S1 4DP, United Kingdom
        \AND
        \name Rashmi Prasad \email rjprasad@linc.cis.upenn.edu \\
        \addr Institute for Research in Cognitive Science, University of Pennsylvania,\\
        3401 Walnut Street, Suite 400A, Philadelphia, PA 19104, USA}

\newcommand{\spg}{SPG}
\newcommand{\spr}{SPR}

\begin{document}

\maketitle

\begin{abstract}
One of the biggest challenges in the development and deployment of
spoken dialogue systems is the design of the spoken language
generation module. This challenge arises from the need for the
generator to adapt to many features of the dialogue domain, user
population, and dialogue context.  A promising approach is trainable
generation, which uses general-purpose linguistic knowledge that is
automatically adapted to the features of interest, such as the
application domain, individual user, or user group.  In this paper we
present and evaluate a trainable sentence planner for providing
restaurant information in the {\sc MATCH} dialogue system.  We show that
trainable sentence planning can produce complex information
presentations whose quality is comparable to the output of a
template-based generator tuned to this domain.  We also show that our
method easily supports adapting the sentence planner to individuals,
and that the individualized sentence planners generally perform better
than models trained and tested on a population of
individuals. Previous work has documented and utilized individual
preferences for content selection, but to our knowledge, these results
provide the first demonstration of individual preferences for sentence
planning operations, affecting the content order, discourse structure and
sentence structure of system responses. Finally, we evaluate the
contribution of different feature sets, and show that, in our
application, n-gram features often do as well as features based on
higher-level linguistic representations.

\end{abstract}

\section{Introduction}
\label{introsec}

One of the most robust findings of studies of human-human dialogue is
that people adapt their interactions to match their conversational
partners' needs and behaviors
\fullcite{Goffman81,BrownLevinson87,PennebakerKing99}.  People adapt the
content of their utterances \fullcite{GarrodAnderson87,LuchokMcCroskey78}.
They choose syntactic structures to match their partners' syntax
\fullcite{LeveltKelter,Braniganetal00,Reitteretal06,svetstent07}, and
adapt their choice of words and referring expressions
\fullcite{CW86,BrennanClark96}.  They also adapt their speaking rate,
amplitude, and clarity of pronunciation
\fullcite{JungersSpeerPalmer02,CoulstonOviattDarves02,FergusonKewley-Port02}.

However, it is beyond the state of the art to reproduce this type of
adaptation in the {\it spoken language generation module} of a
dialogue system, i.e.  the components that handle response generation
and information presentation.  A standard generation system includes
modules for content planning, sentence planning, and surface
realization \fullcite{KKR91,reiter00book}.  A {\bf content planner} takes
as input a communicative goal; it selects content to realize that goal
and organizes that content into a {\em content plan}.  A {\bf sentence
planner} takes as input a content plan.  It decides how the content
is allocated into sentences, how the sentences are ordered, and
which discourse cues to use to express the relationships between
content elements.  It outputs a {\em sentence plan}.  Finally, a {\bf
surface realizer} determines the words and word order for each
sentence in the sentence plan.  It outputs a text or speech {\em
realization} for the original communicative goal.

The findings from human-human dialogue suggest that adaptation could
potentially be useful at any stage of the generation pipeline. Yet to
date, the only work on adaptation to individual users utilizes models
of the user's knowledge, needs, or preferences to adapt the {\it
content} for content planning
\fullcite{JokinenKanto04,Rich79,WahlsterKobsa,ZukermanLitman,CareniniMoore06},
rather than applying models of individual linguistic preferences as to
the {\it form} of the output, as determined by sentence planning or
surface realization.

\begin{figure}[tb]
\begin{center}
\begin{scriptsize}
\begin{tabular}{|p{.1in}|p{3.5in}|c|c|c|} \hline 
Alt & Realization & A & B & AVG \\ \hline 
6 &  Chanpen Thai has the best overall quality among the selected
restaurants since it is a Thai restaurant, with good service, its
price is 24 dollars, and it has good food quality. & 1 & 4 & 2.5 \\ \hline 
7 & Chanpen Thai has the best overall quality among the selected restaurants because it has good service, it has good food quality, it is a Thai restaurant, and its price is 24 dollars. & 2 & 5 & 3.5\\ \hline 
4 & Chanpen Thai has the best overall quality among the selected
restaurants. It has good food quality, with good service, it is a
Thai restaurant, and its price is 24 dollars. & 2 & 4 & 3 \\ \hline 
9 & Chanpen Thai is a Thai restaurant, with good food
quality, its price is 24 dollars, and it has good service. It has
the best overall quality among the selected restaurants. & 2 & 4 & 3 \\ \hline 
5 & Chanpen Thai has the best overall quality
among the selected restaurants. It has good service. It has good
food quality. Its price is 24 dollars, and it is a Thai restaurant.
& 3 & 2 & 2.5 \\ \hline 
3 & Chanpen Thai has the best
overall quality among the selected restaurants. Its price is 24
dollars. It is a Thai restaurant, with good service. It has good
food quality. & 3 & 3 & 3 \\ \hline 
10 & Chanpen Thai has
the best overall quality among the selected restaurants. It has good
food quality. Its price is 24 dollars. It is a Thai restaurant, with
good service. & 3 & 3 & 3 \\ \hline 
2 & Chanpen Thai has
the best overall quality among the selected restaurants. Its price
is 24 dollars, and it is a Thai restaurant. It has good food quality
and good service. & 4 & 4 & 4 \\ \hline 
1 & Chanpen Thai
has the best overall quality among the selected restaurants. This
Thai restaurant has good food quality. Its price is
24 dollars, and it has good service. & 4 & 3 & 3.5   \\ \hline 
8 & Chanpen Thai is a Thai restaurant, with good food quality. It
has good service. Its price is 24 dollars. It has the best overall
quality among the selected restaurants. & 4 & 2 & 3  \\ \hline 
\end{tabular}
\end{scriptsize}
\end{center}
\caption{Some alternative realizations for the content plan in
Figure~\ref{recommend-tplan-fig}, with feedback from Users A and B,
and the mean (AVG) of their feedback
(1$=$worst and 5$=$best).} \label{recommend-alt-fig}
\end{figure}

However, consider the alternative realizations for a restaurant
recommendation in Figure~\ref{recommend-alt-fig}.  Columns A and B
contain human ratings of the quality of the realizations from users A
and B.  The differences in the rating feedback suggest that each user
has different perceptions as to the quality of the potential
realizations.  Data from an experiment collecting feedback from users
A and B, for 20 realizations of 30 different recommendation content
plans (600 examples), shows that the feedback of the two users are
easily distinguished: a paired t-test supports the hypothesis that the
two samples are sampled from distinct distributions ($t=17.4$,
$p<0.001$).  These perceptual differences appear to be more general:
when we examined the user feedback from the evaluation experiment
described by \fullciteA{RambowRogatiWalker01} where 60 users rated the
output of 7 different spoken language generators for 20 content plans,
we again found significant differences in user perceptions of
utterance quality ($F=1.2$, $p<0.002$).  This led us to hypothesize
that individualized sentence planners for dialogue systems might be of
high utility.

In addition to our own studies, we also find evidence in other work
that individual variation is inherent to many aspects of language
generation, including content ordering, referring expression
generation, syntactic choice, lexical choice, and prosody generation.
\begin{itemize}
\item It is common
knowledge that individual authors can be identified from the
linguistic features of their written texts
\fullcite{Madiganetal05,OberlanderBrew00}.
\item An examination of a weather report corpus for five
weather forecasters showed individual differences in lexical choice for
expressing specific weather-related concepts \fullcite{ReiterSripada02}.
\item Rules learned for generating
nominal referring expressions perform better when individual speakers
are provided as a feature to the learning algorithm
\fullcite{JordanWalker2005}, and an experiment evaluating choice of
referring expression shows only 70\% agreement among native speakers
as to the best choice \fullcite{YehMellish97}. \fullciteA{chai2004} show that
there are also individual differences in gesture when generating
multimodal references, and the corpus study of accented pronouns
reported by \fullciteA{kothari07} suggests that accentuation is also
partly determined by individual linguistic style.

\item
Automatic evaluation techniques applied to human-generated reference
outputs for machine translation and automatic summarization perform
better when multiple outputs are provided for comparison
\fullcite{bleu,AnenkovaPassonneauMcKeown07}: this can be attributed to the
large variation in what humans generate given particular content to
express. This is also reflected in the finding that human subjects
produce many different valid content orderings  when asked to order
a specific set of content items to produce the best possible summary
\fullcite{BarzilayElhadadMcKeown02,Lapata2003}.

\end{itemize}

In the past, linguistic variation among individuals was considered a
{\it problem} for generation researchers to work around, rather than a
potential area of study \fullcite{McKeownetal94,Reiter02,Reiteretal03}. In
part, this was due to the hand-crafting of generation components and
resources.  It is impossible to encode by hand, for each individual,
rules for sentence planning and realization.  Furthermore, if domain experts
don't agree on the best way to express a domain concept, how can the
generation dictionary be encoded? It is difficult simply to get
good output that respects all the interacting domain and linguistic constraints even with considerable handcrafting of rules \fullcite{KKR91}.

Modeling individual differences can also be a problem for statistical
methods when learning paradigms are used that assume there is a single
correct output \fullcite{Lapata2003,JordanWalker2005,HardtRambow01} {\it
inter alia}. We believe that the simplest way to deal with the
inherent variability in possible generation outputs is to treat
generation as a ranking problem as we explain below, with techniques
that overgenerate using user or domain-independent rules, and then
filter or rank the possibilities using domain or user-specific corpora
or feedback
\fullcite{LangkildeKnight98,Langkilde2002,BangaloreRambow00,RambowRogatiWalker01}. This
approach has an advantage for dialogue systems because it also affords
joint optimization of the generator and the text-to-speech engine
\fullcite{BulykoOstendorf01,NakatsuWhite06}.  There are many problems in
generation to which ranking models and individualization could be
applied, such as text planning, cue word selection, or referring
expression generation
\fullcite{Mellishetal98,Litman96,Dieugenioetal97,MarciniakStrube04}. However,
only recently has any work in generation acknowledged that there are
individual differences and tried to model them
\fullcite{GuoStent05,MairesseWalker05,Belz05,Lin06}.

This article describes {\sc SPaRKy} (Sentence Planning with Rhetorical
Knowledge), a sentence planner that uses rhetorical relations and
adapts to the user's individual sentence planning
preferences.\footnote{A Java version of {\sc SPaRKy} can be downloaded from
{\tt www.dcs.shef.ac.uk/cogsys/sparky.html}} {\sc SPaRKy} has two
components: a randomized sentence plan generator ({\spg}) that
produces multiple alternative realizations of an information
presentation, and a sentence plan ranker ({\spr}) that is trained
(using human feedback) to rank these alternative realizations (See
Figure~\ref{recommend-alt-fig}).  As mentioned above, previous work has
documented and utilized individual preferences for content selection,
but to our knowledge, our results provide the first demonstration of
individual preferences for sentence planning operations, affecting the
content ordering, discourse structure, sentence structure, and sentence
scope of system responses.  We also show that some of the learned
preferences are domain-specific.

Section~\ref{rel-work-sec} compares our approach and results with
previous work.  Section~\ref{sp-arch-sec} provides an overview of the
{\sc MATCH} system architecture, which can generate dialogue system
responses using either {\sc SPaRKy}, or a domain-specific template-based
generator described and evaluated in previous work
\fullcite{Stentetal02,Walkeretal04}. Sections~\ref{spg-sec},
~\ref{feat-gen-sec} and~\ref{spr-sec} describe {\sc SPaRKy} in detail;
they describe the {\spg}, the automatic generation of features used in
training the {\spr}, and how boosting is used to train the {\spr}.
Sections~\ref{results-sec} and~\ref{qual-results-sec} present both
quantitative and qualitative results:
\begin{enumerate}
\item First, we show that
{\sc SPaRKy} learns to select sentence plans that are significantly better
than a randomly selected sentence plan, and on average less than 10\%
worse than a sentence plan ranked highest by human judges.  We
also show that, in our experiments, simple n-gram
features perform as well as features based on higher-level linguistic
representations.
\item Second, we show
that {\sc SPaRKy}'s {\spg} can produce realizations that are comparable to that of
{\sc MATCH}'s template-based generator, but that there is a gap between
the realization that the {\spr} selects when trained on multiple users and those selected by a human.

\item Third, we show that when {\sc SPaRKy} is trained for particular
individuals, it performs better than when trained on feedback from
multiple individuals. These are the first results suggesting that
individual sentence planning preferences exist, and that they can be
modeled by a trainable generation system. We also show that in most
cases the performance of the individualized {\spr}s are statistically
indistinguishable from {\sc MATCH}'s template-based generator, but for {\sc
compare-2}, User B prefers {\sc SPaRKy}, while for {\sc compare-3}, User A
prefers the template-based generator.

\item Fourth, we show that the differences
in the learned models make sense in terms of previous rule-based
approaches to sentence planning. We analyze the qualitative
differences between the learned group and individual models, and show
that {\sc SPaRKy} learns specific rules about the interaction between content
items and sentence planning operations, and rules that model individual
differences, that would be difficult to capture with a hand-crafted
generator.
\end{enumerate}

We sum up and discuss future work in Section~\ref{discuss-sec}.

\section{Related Work}
\label{rel-work-sec}

We discuss related work on adaptation in generation using the standard
generation architecture which contains modules for content planning
(Section~\ref{ad-cont}), sentence planning (Section~\ref{ad-sp}) and
surface realization (Section~\ref{ad-sr}) \fullcite{KKR91,reiter00book}.

\subsection{Adaptation in Content Planning}
\label{ad-cont}
There has been significant research on the use of user models and
discourse context to adapt the content of information presentations in
dialogue \fullcite{JWW84,JWW86,ChuCarberry95,ZukermanLitman} {\it inter
alia}, but only the user models (not the information presentation strategies) are sensitive to particular
individuals. Several studies have
investigated the use of quantitative models of user preferences in
selection of content for recommendations and comparisons
\fullcite{CareniniMoore06,Walkeretal04,PolifroniWalker06b}, and
\fullciteA{Mooreetal04} use such models for referring expression
generation, sentence planning and some surface
realization. \fullciteA{Elhadadetal2005} applied group models (physician,
lay person) and individual user models to the task of summarizing
medical information.

\fullciteA{McCoy89} used context information to design
helpful system-generated corrections.  Other work has looked at the use of statistical
techniques for adapting content selection and content ordering methods
to particular domains
\fullcite{BarzilayElhadadMcKeown02,DuboueMcKeown03,Lapata2003}, but not to
individual users.

\subsection{Adaptation in Sentence Planning}
\label{ad-sp}

The first trainable sentence planner was SPoT, a precursor to {\sc SPaRKy}
that output information gathering utterances in the travel domain
\fullcite{WRR02}. Evaluations of SPoT demonstrated that it performed as
well as a template-based generator developed for the travel domain and
field-tested in the DARPA Communicator evaluations
\fullcite{RambowRogatiWalker01,Walkeretal02b}.  Information gathering
utterances are considerably simpler than information presentations:
they do not usually exhibit any complexities in rhetorical structure, and
there is little interaction between domain-specific content items and
sentence structures.  Thus the SPoT generator did not produce
utterances with variation in rhetorical structure; it learned to
optimize speech-act ordering and sentence structure choices, but it
did not adapt to individuals.

\subsection{Adaptation in Surface Realization}
\label{ad-sr} Work on adaptation in surface realization has mainly focused on decisions such as lexical and syntactic choice,
using models of a target text, but not individual text models,
although recent research has also shown that n-gram models trained on
user-specific corpora can adapt generators to reproduce individualized
lexical and syntactic choices
\fullcite{Lin06,Belz05}. \fullciteA{PaivaEvans2004} present a technique for
training a generator by learning the relationship between particular
generation decisions and text variables that can be measured in the
output corpus. This technique was applied to generator decisions such
as the form of referring expression and syntactic structure, and was
used to capture stylistic, rather than individual, differences.
\fullciteA{GuptaStent05} use discourse context and speaker knowledge for
referring expression generation in dialogue.

User models have also been used to adapt surface
realization.  The approach of learning a ranking from user feedback
has been applied to multimedia presentation planning \fullcite{StentGuo05}
and to the joint optimization of the syntactic realizer and the
text-to-speech engine \fullcite{NakatsuWhite06}. This work does not look
at individual differences.

Research has also focused on other factors that affect stylistic
variation -- how realization choices reflect personality, politeness, emotion or
domain specific style
\fullcite{Hovy87,DiMarcoFoster97,WCW97,AndreRist,bouayadagha2000ica,FleishmanHovy2002,Piwek03,PorayskaMellish04,Isardetal06,guptawalkerromano07,mairessewalker07}. None of this work has attempted to reproduce
individual stylistic variation.


\newpage

\section{Overview of {\sc MATCH}'s Spoken Language Generator}
\label{sp-arch-sec}

\begin{figure}[htb]
  \centerline{\psfig{figure=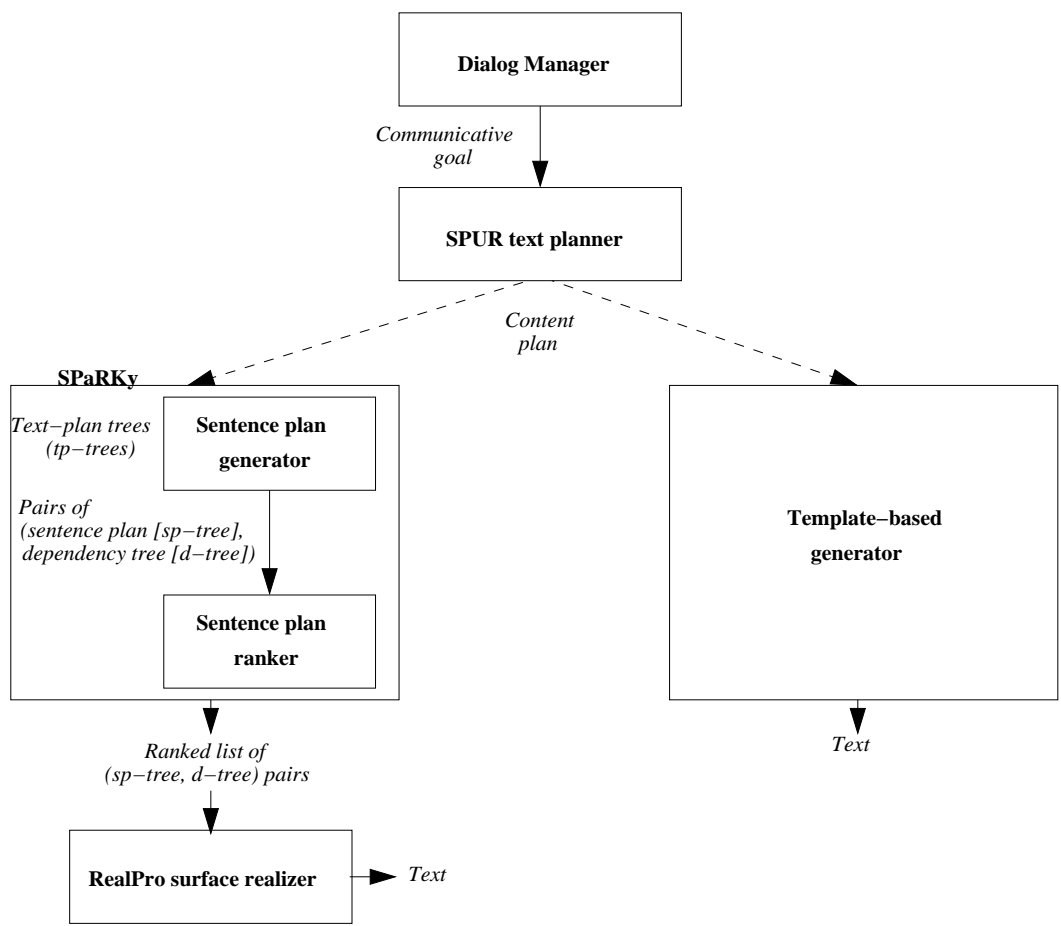,width=5.5in}}
  \caption{Architecture of {\sc MATCH}'s Spoken Language Generator.}
  \label{arch-fig}
\end{figure}

\begin{figure}[htb]
\centering
\begin{scriptsize}
\begin{tabular}{|p{0.8in}|p{0.6in}|p{3.0in}|c|} \hline
Strategy & System & Realization & AVG \\ \hline \hline
{\sc recommend}  &
Template & Caffe Cielo has the best overall value among the selected
restaurants. Caffe Cielo has good decor and good service. It's an
Italian restaurant.
 &   4 \\ \hline
{\sc recommend}  & {\sc SPaRKy} & Caffe Cielo, which is an Italian
restaurant, with good decor and good service, has the best overall
quality among the selected restaurants. & 4 \\ \hline \hline
{\sc compare-2}  & Template & Caffe Buon Gusto's an Italian restaurant. On
the other hand, John's Pizzeria's an Italian, Pizza restaurant.
 &  2 \\  \hline
{\sc compare-2}  & {\sc SPaRKy} & Caffe Buon Gusto is an Italian restaurant,
and John's Pizzeria is an Italian , Pizza restaurant. & 4
\\ \hline \hline
{\sc compare-3} & Template & Among the selected restaurants, the following offer exceptional overall value. Uguale's
price is 33 dollars.  It has good decor and very good service.  It's a French, Italian restaurant.  Da Andrea's price is 28
dollars.  It has good decor and very good service.  It's an Italian restaurant.  John's Pizzeria's price is 20 dollars.  It has
mediocre decor and decent service. It's an Italian, Pizza restaurant. & 4.5 \\ \hline {\sc compare-3} & {\sc SPaRKy} & Da Andrea,
Uguale, and John's Pizzeria offer exceptional
 value among the selected restaurants. Da Andrea is an Italian
 restaurant, with very good service, it has good decor, and its price
 is 28 dollars. John's Pizzeria is an Italian , Pizza restaurant. It
 has decent service. It has mediocre decor. Its price is 20
 dollars. Uguale is a French, Italian restaurant, with very good
 service. It has good decor, and its price is 33 dollars. & 4 \\ \hline
\end{tabular}
\end{scriptsize}
\caption{Template outputs and a sample {\sc SPaRKy} output for each
dialogue strategy.  AVG = Averaged score of two human users.}
\label{templatevssparky}
\end{figure}

{\sc MATCH} (Multimodal Access To City Help) is a multimodal dialogue system
for finding restaurants and entertainment options in New York City
\fullcite{Johnstonetal02}.  Information presentations in {\sc MATCH} include
route descriptions, as well as user-tailored {\it recommendations} and
{\it comparisons} of restaurants. Figure~\ref{arch-fig} shows {\sc MATCH}'s
architecture for spoken language generation (SLG).  The content
planning module is the {\sc SPUR} text planner (Section~\ref{spur-sec})
\fullcite{Walkeretal04}.  There are two modules for producing text or
spoken dialogue responses from {\sc SPUR}'s output: a highly
engineered domain-specific template-based realizer
(Section~\ref{template-sec}); and the {\sc SPaRKy} sentence
planner followed by the RealPro surface realizer \fullcite{LavoieRambow97}
(Section~\ref{sparky-sec}).  Example template-based and {\sc SPaRKy} outputs
for each dialogue strategy are in Figure~\ref{templatevssparky}.
Both {\sc SPUR} and {\sc SPaRKy} are trainable, and produce different output
depending on the user and discourse context.

\subsection{{\sc SPUR}}
\label{spur-sec}
The input to {\sc SPUR} is a high-level {\bf communicative goal} from the
{\sc MATCH} dialogue manager and its output is a {\bf content plan} for a
recommendation or comparison.
{\sc SPUR} selects and organizes the content to be
communicated based on the communicative goal, a
conciseness parameter, and a decision-theoretic user model.
It produces {\em targeted} recommendations and comparisons: the
restaurants mentioned and the
attributes selected for each restaurant are those the user
model predicts the user will want to know about. Thus {\sc SPUR}
can produce a wide variety of content plans.

Figure~\ref{recommend-tplan-fig} shows a sample content plan for a
recommendation.  This content plan gives rise to the
alternate realizations for recommendations for Chanpen Thai in
Figure~\ref{recommend-alt-fig}. Following a bottom-up approach to
text-planning \fullcite{Marcu97,Mellishetal98}, each content plan consists
of a set of {\it assertions} that must be communicated to the user and
a set of {\it rhetorical relations} that hold between those assertions
that may be communicated as well. Each rhetorical relation designates
one or more facts as the {\it nuclei} of the relation, i.e. the main
point, and the other facts as {\it satellites}, i.e. the supplementary
facts \fullcite{MannThompson87}. Three rhetorical relations
\fullcite{MannThompson87} are used by {\sc SPUR}: the {\sc justify} relation for the
recommendation strategy, and the {\sc contrast} and {\sc elaboration}
relations for the comparison strategies.  The relations in
Figure~\ref{recommend-tplan-fig} specify that the nucleus (1) is the
{\it claim} being made in the recommendation, and that the satellites
(assertions 2 to 5) provide justifying evidence for the claim.

\begin{figure}[htb]
\begin{center}
\begin{tabular}{|p{0.45in}p{4.35in}|}
\hline
 relations: &  justify(nuc:1, sat:2); justify (nuc:1, sat:3 ); justify(nuc:1, sat:4); justify(nuc:1, sat:5) \\


 content: & 1. assert(best (Chanpen Thai)) \\
          & 2. assert(is (Chanpen Tai, cuisine(Thai))) \\
          & 3. assert(has-att(Chanpen Thai, food-quality(good)))\\
          & 4. assert(has-att(Chanpen Thai, service(good))) \\
          & 5. assert(is (Chanpen Thai, price(24 dollars))) \\

\hline
\end{tabular}
\caption{A content plan for a recommendation.}
\label{recommend-tplan-fig}
\end{center}
\end{figure}

\subsection{Template-Based Generator}
\label{template-sec}

In order to produce utterances from the content plans produced by
{\sc SPUR}, we first implemented and evaluated a template-based generator
for {\sc MATCH} \fullcite{Stentetal02,Walkeretal04}.
The template-based generator was designed to make it possible to
evaluate algorithms for user-specific content selection based on
{\sc SPUR}'s decision-theoretic user model. It performs sentence planning,
including some discourse cue insertion, clause combining and referring
expression generation.  It produces one high quality output for any
content plan for our three dialogue strategies: {\sc recommend}, {\sc
compare-2} and {\sc compare-3}. Recommendations and comparisons are
one form of {\it evaluative argument}, so its realization strategies
are based on guidelines from argumentation theory for producing
effective evaluative arguments, as summarized by
\fullciteA{careninimoore00b}.  Because the templates are highly tailored to
this domain, the template-based generator can be expected to perform
well in comparison to {\sc SPaRKy}.

Following the argumentation guidelines, the template-based generator
realizes recommendations with the nucleus ordered first, followed by
the satellites. The satellites are ordered to maximize the opportunity
for aggregation. To produce the most concise recommendations given the
content to be communicated, phrases with identical verbs and subjects
are grouped, so that lists and coordination can be used to aggregrate
the assertions about the subject.  Figure~\ref{concise-recommends-fig}
provides examples of aggregration  as the number
of assertions varies according to {\sc SPUR}'s conciseness parameter
(Z-value).

\begin{figure}
\begin{center}
\begin{scriptsize}
\begin{tabular}{|c|p{4.5in}|} \hline
Z  & Output \\ \hline \hline 1.5  & Komodo has the best overall value among the selected restaurants. Komodo's a Japanese, Latin
American restaurant. \\ \hline 
0.7 & Komodo has the best overall value among the selected restaurants. Komodo's a Japanese, Latin American restaurant.  \\ \hline 
0.3 & Komodo has the best overall value among the selected restaurants. Komodo's  price is \$29. It's a Japanese, Latin American restaurant.  \\ \hline 
-0.5 & Komodo has the best overall value among the selected restaurants. Komodo's  price is \$29 and it has very good service. It's a Japanese, Latin American restaurant.  \\ \hline 
-0.7 & Komodo has the best overall value among the selected restaurants. Komodo's price is \$29 and it has very good service and very good food quality. It's a Japanese, Latin American restaurant.   \\ \hline 
-1.5  &  Komodo has the best overall value among the selected restaurants. Komodo's price is \$29 and it has very good service, very good food quality and good decor. It's a Japanese, Latin American restaurant. \\ \hline 
\end{tabular}
\end{scriptsize}
\end{center}
\caption{Recommendations for the East Village Japanese Task, for
different settings of the conciseness parameter Z. \label{concise-recommends-fig} }
\end{figure}


\begin{figure}[h]
\begin{center}
\begin{tabular}{|p{0.5in}p{4.5in}|} \hline
 strategy: & compare3 \\
 items: &  Above, Carmine's \\ \hline
  relations: &  elaboration(nuc:1,sat:2); elaboration(nuc:1,sat:3); elaboration(nuc:1,sat:4); elaboration(nuc:1,sat:5); elaboration(nuc:1,sat:6); elaboration(nuc:1,sat:7); contrast(nuc:2,nuc:3); contrast(nuc:4,nuc:5); contrast(nuc:6,nuc:7)\\
  content: & 1. assert(exceptional(Above,Carmine's)) \\
           & 2. assert(has-att(Above, decor(good))) \\
           & 3. assert(has-att(Carmine's, decor(decent))) \\
           & 4. assert(has-att(Above, service(good))) \\
           & 5. assert(has-att(Carmine's, service(good))) \\
           & 6. assert(has-att(Above, cuisine(New American))) \\
           & 7. assert(has-att(Carmine's, cuisine(Italian))) \\ \hline
\end{tabular}
\caption{A content plan for a comparison.}
\label{midtownwest-kashi-compare3-tplan-fig}
\end{center}
\end{figure}

The realization template for comparisons focuses on communicating both
the {\it elaboration} and the {\it contrast} relations.
Figure~\ref{midtownwest-kashi-compare3-tplan-fig} contains a content
plan for comparisons. The nucleus is the assertion (1) that Above and
Carmine's are exceptional restaurants.  The satellites (assertions 2
to 7 representing the selected attributes for each restaurant) {\it
elaborate} on the claim in the nucleus (assertion 1).  {\it Contrast}
relations hold between assertions 2 and 3, between 4 and 5, and
between 6 and 7.  One way to communicate the {\it elaboration}
relation is to structure the comparison so that all the satellites are
grouped together, following the nucleus. To communicate the
{\it contrast} relation, the satellites are produced in a fixed
order, with a parallel structure maintained across options
\fullcite{Prevost95,Prince85}. The satellites are initially ordered in
terms of their evidential strength, but then are reordered to allow
for aggregation.  Figure~\ref{concise-compares-fig} illustrates aggregation
for comparisons with varying numbers of
assertions.

\begin{figure}[h]
\begin{center}
\begin{scriptsize}
\begin{tabular}{|c|p{4.5in}|} \hline
Z  & Output \\ \hline \hline
1.5  & Among the selected restaurants, the following offer exceptional overall value. Komodo has very good service. \\ \hline 
0.7  & Among the selected restaurants, the following offer exceptional overall value. Komodo has very good service and good decor.  \\ \hline 
0.3 &Among the selected restaurants, the following offer exceptional overall value. Komodo's price is \$29.  It has very good food quality, very good service  and  good decor. Takahachi's price is \$27.  It has very good food quality, good service   and  decent decor. \\ \hline 
-0.5 & Among the selected restaurants, the following offer exceptional overall value. Komodo's price is \$29.  It has very good food quality, very good service  and  good decor. Takahachi's price is \$27.  It has very good food quality, good service   and  decent decor. Japonica's price is\$37.  It has excellent food quality, good service   and  decent decor  \\ \hline 
-0.7  & Among the selected restaurants, the following offer exceptional overall value. Komodo's price is \$29.  It has very good food quality, very good service  and  good decor. Takahachi's price is \$27.  It has very good food quality, good service   and  decent decor. Japonica's price is \$37.  It has excellent food quality, good service   and  decent decor. Shabu-Tatsu's price is \$31.  It has very good food quality, good service   and  decent decor. \\ \hline 
-1.5 & Among the selected restaurants, the following offer exceptional overall value. Komodo's price is \$29.  It has very good food quality, very good service  and  good decor. Takahachi's price is \$27.  It has very good food quality, good service   and  decent decor. Japonica's price is \$37.  It has excellent food quality, good service   and  decent decor. Shabu-Tatsu's price is \$31.  It has very good food quality, good service   and  decent decor. Bond Street's price is \$51.  It has excellent food quality, good service   and  very good decor. Dojo's price is \$14.  It has decent food quality, mediocre service   and  mediocre decor. \\ \hline  
\end{tabular}
\end{scriptsize}
\end{center}
\caption{Comparisons for the East Village Japanese Task, for
different settings of the conciseness parameter Z. \label{concise-compares-fig}}

\end{figure}

\subsection{{\sc SPaRKy}}
\label{sparky-sec}

Like the template-based generator, {\sc SPaRKy} takes as input any of the
content plans produced by {\sc SPUR}. Figure~\ref{arch-fig} shows that
{\sc SPaRKy} has two modules: the sentence plan generator (SPG), and the
sentence plan ranker (SPR).  The {\spg} uses a set of clause-combining
operations (Figure~\ref{syn-ops-examples-fig}); it produces a
large set of alternative realizations of an input content plan (See
Figure~\ref{recommend-alt-fig}).  The {\spr} ranks
the alternative realizations using a model learned from users'
ratings of a training set of content plans.  The {\spg} is described
in Section~\ref{spg-sec}.  The features used to train the {\spr} are
described in Section~\ref{feat-gen-sec}; the procedure for training
the {\spr} is described in Section~\ref{spr-sec}.

Because {\sc SPaRKy} is trained using user feedback, rather than being
handcrafted, it can be trained to be an individualized spoken language
generator. As discussed above, the feedback from the two users in
Figure~\ref{recommend-alt-fig} suggests that each user has different
perceptions as to the quality of the potential realizations.  A
significant part of Sections~\ref{results-sec} and
~\ref{qual-results-sec} are dedicated to examining the differences
between a model trained on averaged feedback, shown as AVG in
Figure~\ref{recommend-alt-fig}, and those trained on individual
feedback from users A and B.




\section{Sentence Plan Generation}
\label{spg-sec}

\begin{figure}[h]
  \centering

\begin{scriptsize}
\begin{tabular}{|p{.1in}|p{4.0in}|c|c|c|} \hline
 Alt & Realization & A & B & AVG  \\ \hline \hline
11 & Above and Carmine's offer exceptional value among the selected restaurants. Above, which is a New American restaurant, with good decor, has good service. Carmine's, which is an Italian restaurant, with good service, has decent decor.&2 & 2 &  2 \\ \hline %

12 & Above and Carmine's offer exceptional value among the selected restaurants. Above has good decor, and Carmine's has decent decor. Above and Carmine's have good service. Above is a New American restaurant. On the other hand, Carmine's is an Italian restaurant. &3 &2 & 2.5  \\ \hline %

13 & Above and Carmine's offer exceptional value among the selected restaurants. Above is a New American restaurant. It has good decor. It has good service. Carmine's, which is an Italian restaurant, has decent decor and good service. & 3& 3& 3   \\ \hline %

14 & Above and Carmine's offer exceptional value among the selected restaurants. Above has good decor while Carmine's has decent decor, and Above and Carmine's have good service. Above is a New American restaurant while Carmine's is an Italian restaurant. &4 &5 & 4.5   \\ \hline %


20 & Above and Carmine's offer exceptional value among the selected restaurants. Carmine's has decent decor but Above has good decor, and Carmine's and Above have good service. Carmine's is an Italian restaurant. Above, however, is a New American restaurant. &2 & 3 & 2.5   \\ \hline %

25 & Above and Carmine's offer exceptional value among the
  selected restaurants. Above has good decor. Carmine's is an Italian
  restaurant. Above has good service. Carmine's has decent decor.
  Above is a New American restaurant. Carmine's has good service. & NR & NR & NR \\ \hline
\end{tabular}
\end{scriptsize}
\caption{Some alternative realizations for the {\sc compare-3} plan in
Figure ~\ref{midtownwest-kashi-compare3-tplan-fig},
with feedback from Users A and B,
and the mean (AVG) of their feedback
(1$=$worst and 5$=$best). NR = Not generated or ranked.} \label{compare3-alt-fig}
\end{figure}

\begin{figure}[htb]
\begin{center}
\includegraphics[width=20pc]{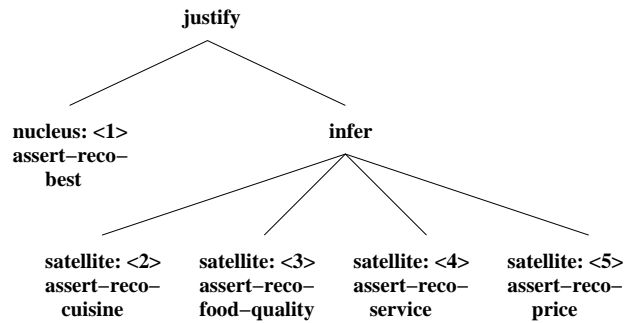}

\caption{A tp-tree for the plan of Figure~\ref{recommend-tplan-fig},
used to generate Alternatives 1, 3, 4, 5, 6, 7 and 10 in
Figure~\ref{recommend-alt-fig}.} \label{alt6-tplantree}
\end{center}
\end{figure}

\begin{figure*}[htb]
\centering
  \centerline{\psfig{figure=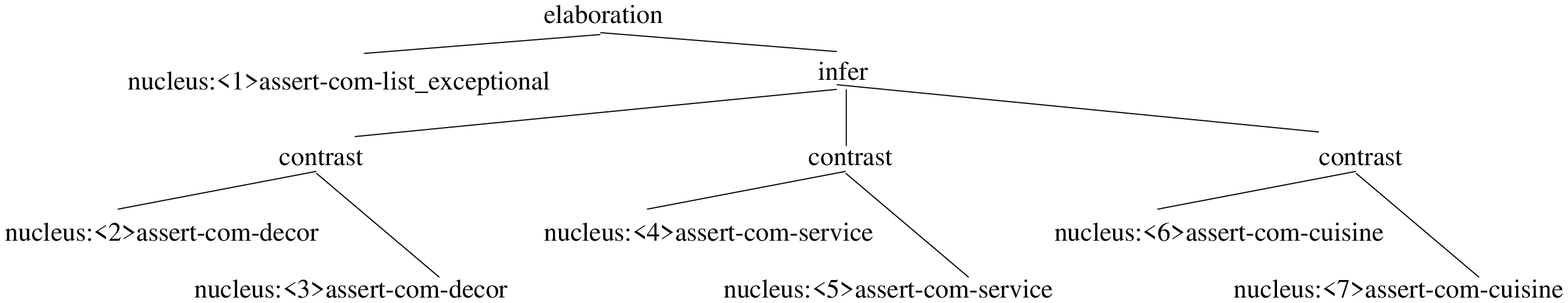,height=1.35in,width=6.05in}}
\vspace{2em}
  \centerline{\psfig{figure=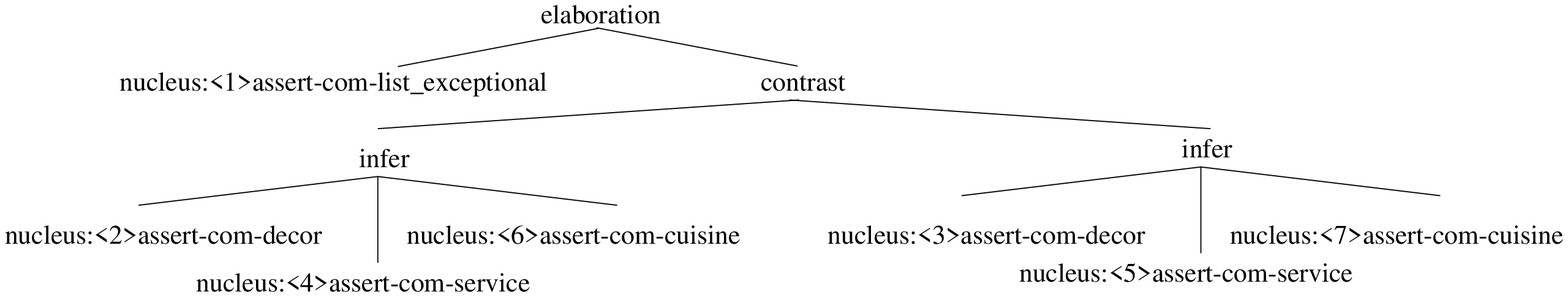,height=1.3in,width=6.05in}}
\caption{Tp-trees for the comparisons shown as alternatives 12 and 14 (top) and alternatives 11 and 13 (bottom)
in
  Figure~\ref{compare3-alt-fig}.}
\label{midtownwest-KashiVishwanath_k0.7_compare3-rst_transformed_plans.out2}
\end{figure*}

The input to {\sc SPaRKy}'s {\spg} is a {\bf content plan} from {\sc SPUR}.
Content plans for a sample recommendation and comparison were
in Figure~\ref{recommend-tplan-fig} and
Figure~\ref{midtownwest-kashi-compare3-tplan-fig}.
Figure~\ref{recommend-alt-fig} shows alternative {\sc SPaRKy} realizations for the recommendation in Figure~\ref{recommend-tplan-fig}, while Figure~\ref{compare3-alt-fig} shows alternative {\sc SPaRKy} realizations
for the comparison in Figure
~\ref{midtownwest-kashi-compare3-tplan-fig}.  Content plans specify
which assertions to include in an information presentation, and the
rhetorical relations holding between them, but not the
order of assertions or how to express the rhetorical relations
between them. This task is known as {\em discourse planning}.  The
{\spg} has two stages of processing; first it does discourse planning,
and then it does sentence planning.

\subsection{Discourse Planning}

Discourse planning algorithms can be characterized as:
schema-based \fullcite{McKeown85,KKR91}; top-down algorithms using
plan operators \fullcite{MooreParis93}; or bottom-up approaches
that use, for example, constraint satisfaction algorithms
\fullcite{Marcu96,Marcu97} or genetic algorithms \fullcite{Mellishetal98}.  In
{\sc SPaRKy}, the {\spg} takes a bottom-up approach to discourse planning
using principles from Centering Theory \fullcite{GJW95}. Content items are
grouped because they talk about the same thing, but the linear order
between and among the groupings is left unspecified.  The centering
constraints have the result that Alt-25 in
Figure~\ref{compare3-alt-fig}, which repeatedly changes the discourse
center, are never generated.

The discourse planning stage produces one or more text-plan trees
({\bf tp-trees}).  A tp-tree for the {\sc recommend} plan in
Figure~\ref{recommend-tplan-fig} is in Figure~\ref{alt6-tplantree},
and tp-trees for the {\sc compare-3} plan in
Figure~\ref{midtownwest-kashi-compare3-tplan-fig} are in
Figure~\ref{midtownwest-KashiVishwanath_k0.7_compare3-rst_transformed_plans.out2}.
In a tp-tree, each leaf represents a single assertion and is labeled
with a speech act.  Interior nodes are labeled with rhetorical
relations.  In addition to the rhetorical relations in the content
plan, the {\spg} uses the relation {\sc infer} for combinations of
speech acts for which there is no rhetorical relation expressed in the
content plan \fullcite{Marcu97}.  The {\it infer} relation is
similar to the {\it joint} relation in RST; it joins
multiple satellites in a mononuclear relation or the nuclei in a
multinuclear relation.

Each simple assertion, or leaf, in a tp-tree is associated with one or
more syntactic realizations ({\bf d-trees}), using a dependency tree
representation, called DSyntS (Figure~\ref{dictionary}) \fullcite{Melcuk88,LavoieRambow97}. The association between the
simple assertions and any potential d-trees specifying their syntactic
realizations is specified in a hand-crafted generation dictionary.
Leaves of some d-trees in the generation dictionary are variables,
which are instantiated from the content plan, e.g. {\it Thai} replaces
a cuisine type variable.

\begin{figure*}[ht]
\centering
\begin{scriptsize}
\begin{tabular}{ll}\hline
assert-com-cuisine & BE3 [class:verb ]\\
& \hspace{0.05in} (\\
& \hspace{0.1in} I Chanpen\_Thai [number:sg class:proper\_noun article:no-art person:3rd ]\\
& \hspace{0.1in} II restaurant [class:common\_noun article:indef ]\\
& \hspace{0.15in} ( \\
& \hspace{0.2in} Thai [class:adjective ]\\
& \hspace{0.15in} )\\
& \hspace{0.05in} )\\ \hline
assert-com-food\_quality & HAVE1 [class:verb ] \\
& \hspace{0.05in} (\\
& \hspace{0.1in} I Chanpen\_Thai [number:sg class:proper\_noun article:no-art person:3rd ]\\
& \hspace{0.1in} II quality [class:common\_noun article:no-art ]\\
& \hspace{0.15in} ( \\
& \hspace{0.2in} ATTR good [class:adjective ]\\
& \hspace{0.2in} ATTR food [class:common\_noun ]\\
& \hspace{0.15in} )\\
& \hspace{0.05in} )\\ \hline
\hline
\end{tabular}
\end{scriptsize}
\caption{Example d-trees from the generation dictionary used by the \spg\label{dictionary}.}
\end{figure*}

\begin{figure*}[ht]

\begin{scriptsize}
\begin{tabular}{|p{0.55in}p{0.6in}p{1.7in}p{0.7in}p{0.65in}p{1.0in}|} \hline
{\bf Operation} & {\bf Rel} & {\bf Description} & {\bf Sample 1st
arg} & {\bf Sample 2nd arg} &{\bf Result}\\ \hline

M{\sc erge} & {\sc infer} or {\sc contrast} & Two clauses can be
combined if they have identical matrix verbs and identical arguments
and adjuncts except one. The non-identical arguments are
coordinated. & Chanpen Thai has good service. & Chanpen Thai has
good food quality. & Chanpen Thai has good
service and good food quality.\\
\hline

W{\sc ith-reduction} & {\sc justify} or {\sc infer} & Two clauses
with identical subject arguments can be identified if one of the
clauses has a {\sc have}-possession matrix verb. The possession
clause undergoes {\it with}-participial clause formation and is
attached to the non-reduced clause.\footnotemark[4] & Chanpen Thai is a Thai restaurant. &
Chanpen Thai has good food quality. & Chanpen Thai is a Thai
restaurant, with good food quality.\\ \hline

R{\sc elative-clause} & {\sc justify} or {\sc infer} & Two clauses
with an identical subject can be identified. One clause is attached
to the subject of the other clause as a relative clause.\footnotemark[5] & Chanpen Thai has the best overall quality among the
selected restaurants. & Chanpen Thai is located in Midtown West. &
Chanpen Thai, which is located in Midtown West, has the best overall
quality among the selected restaurants.\\ \hline

C{\sc ue-word-conjunction} &
{\sc justify}, {\sc infer} or {\sc contrast}
& Two clauses are conjoined with a cue word
(coordinating or subordinating conjunction).
The order of the arguments of the connective is determined by the
order of the nucleus (N) and the satellite (S), yielding
two distinct operations, {\sc cue-word-conjunction-ns}
and {\sc cue-word-conjunction-sn}.
&  Chanpen Thai has
the best overall quality among the selected restaurants. & Chanpen
Thai is a Thai restaurant, with good service. & Chanpen Thai has the
best overall quality among the selected restaurants, since it is a
Thai restaurant, with good service.
\\ \hline

C{\sc ue-word-insertion} ({\it on the other hand}) & {\sc contrast}
& {\sc cue-word insertion} combines clauses by inserting a cue word
at the start of the second clause ({\it Carmine's is an Italian
restaurant. HOWEVER, Above is a New American restaurant}), resulting
in two separate sentences.  & Penang has very good decor. &
Baluchi's has mediocre decor. & Penang has very good decor. On the
other hand, Baluchi's has mediocre
decor.\\
\hline

P{\sc eriod} & {\sc justify}, {\sc contrast}, {\sc infer} or {\sc elaboration} &
Two clauses are joined by a period. & Chanpen Thai is a Thai
restaurant, with good food quality. & Chanpen Thai has good service.
& Chanpen Thai is a
Thai restaurant, with good food quality. It has good service.\\
\hline
\end{tabular}
\caption{Clause combining operations and examples.}
\label{syn-ops-examples-fig}
\end{scriptsize}
\end{figure*}

\subsection{Sentence Planning}

During sentence planning, the {\spg} assigns assertions to sentences,
orders the sentences, inserts discourse cues, and performs referring
expression generation.  It uses a set of clause-combining operations
that operate on tp-trees and incrementally transform the elementary
d-trees associated with their leaves into a single lexico-structural
representation.  The output from this process is two parallel
structures: (1) a sentence plan tree ({\bf sp-tree}), a binary tree
with leaves labeled with the assertions from the input tp-tree, and
interior nodes labeled with clause-combining operations; and (2) one
or more {\bf d-trees} which reflect parallel operations on the
predicate-argument representations.

The clause-combining operations are general operations similar to
aggregation operations used in other research
\fullcite{RambowKorelsky92,danlos2000}. The operations and examples of
their use are given in Figure~\ref{syn-ops-examples-fig}.  They are
applied in a bottom-up left-to-right fashion, with the choice of
operation constrained by the rhetorical relation holding between the
assertions to be combined \fullcite{scott90}, as specified in
Figure~\ref{syn-ops-examples-fig}.

In addition to ordering assertions, a clause-combining operation may
insert cue words between assertions.  Figure~\ref{fig:rst-op-mapping}
gives the list of cue words used by the {\spg}. The choice of cue-word
is determined by the type of rhetorical relation\footnote{An alternative
approach is for the cue-word to impose a constraint on the rhetorical
relation that must hold \fullcite{webber99little-trees,forbesetal03}.}.



\begin{figure}[htb]
\centering
  \footnotesize
  \begin{tabular}{|l|p{4.5in}|} \hline
   {\bf RST relation} &    {\bf Aggregation operator}\\ \hline \hline
    {\sc justify}   &       {\sc with-reduction}, {\sc relative-clause}, {\sc cue-word conj.} {\it because}, {\sc cue-word conj.} {\it since}, {\sc period} \\ \hline
   {\sc contrast}  &       {\sc merge}, {\sc cue-word insert.} {\it however}, {\sc cue-word conj.} {\it while}, {\sc cue-word conj.} {\it and}, {\sc cue-word conj.} {\it but}, {\sc cue-word insert.} {\it on the other hand}, {\sc period} \\ \hline
   {\sc infer}     &       {\sc merge}, {\sc cue-word conj.} {\it and}, {\sc period} \\ \hline
    {\sc elaboration}&      {\sc period} \\ \hline
  \end{tabular}
  \caption{\label{fig:rst-op-mapping} RST relation constraints on aggregation
    operators.}
\end{figure}

\begin{figure}[htb]
\centering
  \footnotesize
  \begin{tabular}{|p{4.5in}|c|} \hline
   {\bf Aggregation operator} & {\bf Probability}\\ \hline \hline
         {\sc merge}, {\sc with-reduction}, {\sc relative-clause} & 0.80\\ \hline
   {\sc cue-word conj.} {\it because}, {\sc cue-word conj.} {\it since}, {\sc cue-word conj.} {\it while}, {\sc cue-word conj.} {\it and}, {\sc cue-word conj.} {\it but} & 0.10 \\ \hline
     {\sc cue-word insert.} {\it however}, {\sc cue-word insert.} {\it on the other hand} & 0.09\\ \hline
     {\sc period} & 0.01\\ \hline
  \end{tabular}
  \caption{\label{fig:aggreg-prob}Probability distribution of aggregation operators. The final operation is randomly chosen from the selected set with a uniform distribution.}
\end{figure}

The {\spg} generates a random sample of possible sp-trees for each
tp-tree, up to a pre-specified number of sp-trees, by randomly
selecting among the clause-combining operations according to a
probability distribution that favors preferred operations. Table~\ref{fig:aggreg-prob} shows the
probability distribution used in our experiments, which is hand-crafted based on assumed preferences
for operations such as {\sc merge}, {\sc relative-clause} and {\sc
with-reduction}, and is one way in which some knowledge can be
injected into the random process to bias it towards producing higher
quality sentence plans.\footnote{This
probability distribution could be learned from a corpus
\fullcite{Marcu97,rashmi-gen-disc-treebank-paper}.}

\stepcounter{footnote}
\stepcounter{footnote}

\footnotetext[4]{If an {\sc infer}
relation holds and both clauses contain the {\sc have} possession
predicate, the second clause is arbitrarily selected for reduction.
If a {\sc justify} relation holds, it is the satellite of the RST
relation that always undergoes reduction, if the syntactic
constraints are satisfied.}

\footnotetext[5]{If
an {\sc infer} relation holds, any clause is arbitrarily selected
for reduction. If a {\sc justify} relation holds, the clause that
undergoes relative clause formation is the satellite clause. %
This is motivated by the fact that relative clause formation is
generally seen to occur when the modifying relative clause provides
additional information about the noun it modifies, but where the
additional/elaborated information does not have the same {\it
  informational} status as the information in the main
clause.}



\begin{figure}[htb]
\begin{center}
\includegraphics[width=3.3in]{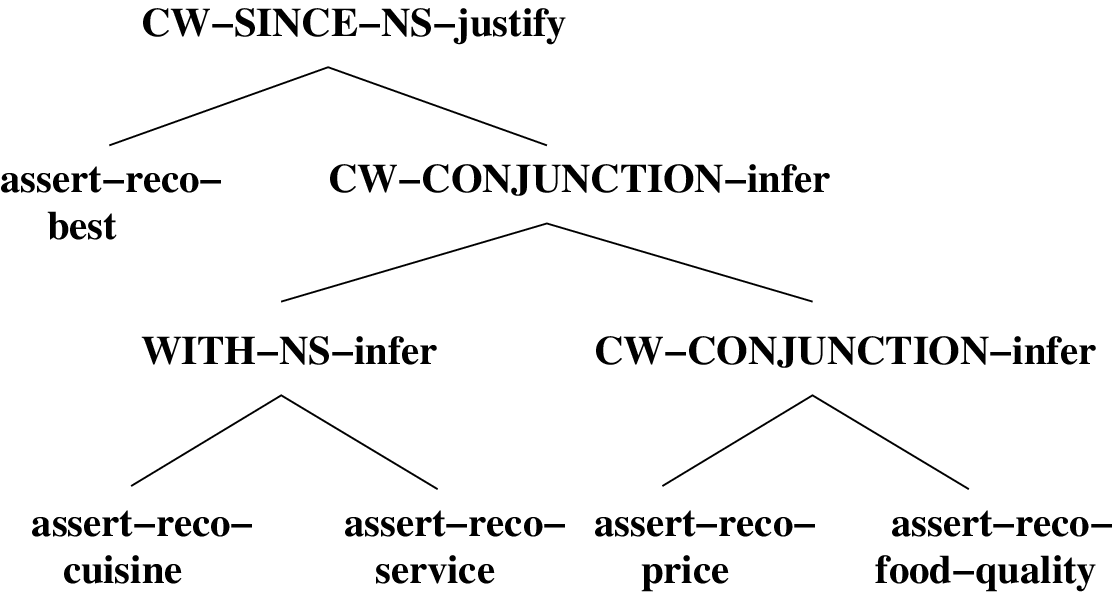}
\caption{Sentence Plan Tree (SP-tree) for Alternative 6 of
Figure~\ref{recommend-alt-fig}.} \label{alt6-sptree}
\end{center}
\end{figure}

\begin{figure}[htb]
\begin{center}
\includegraphics[width=3.3in]{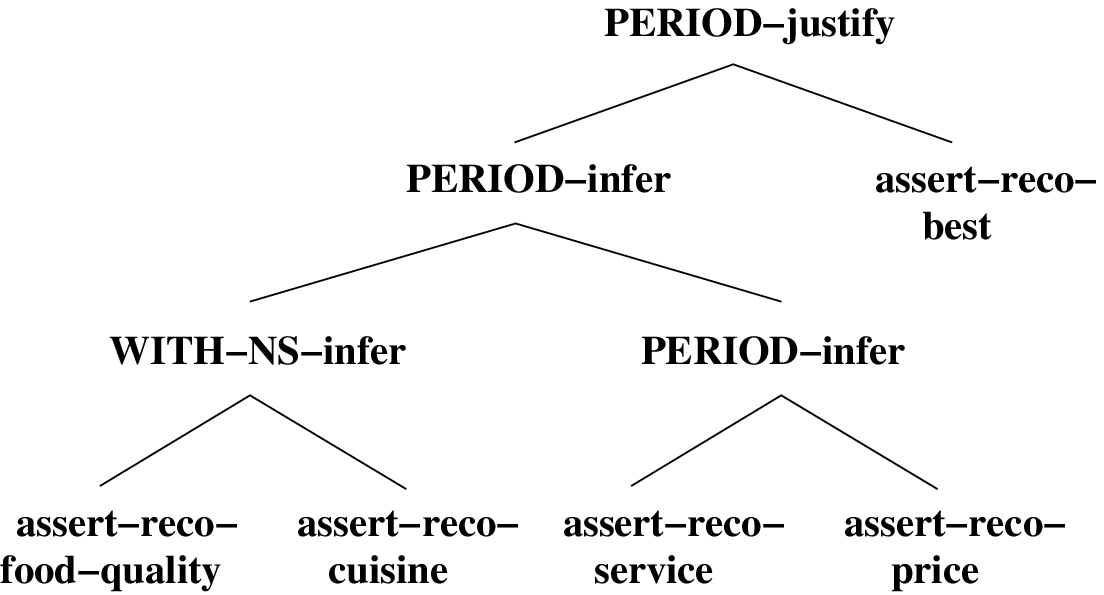}
\caption{Sentence Plan Tree (SP-tree) for Alternative 8 of
Figure~\ref{recommend-alt-fig}.} \label{alt8-sptree}
\end{center}
\end{figure}


The {\spg} handles referring expression generation by converting a
proper name to a pronoun when the same proper name appears in the
previous utterance. Referring expression generation rules are applied
locally, across adjacent utterances, rather than globally across the
entire presentation at once \fullcite{BFP87}.  Referring expressions are
manipulated in the d-trees, either intrasententially during the
incremental creation of the sp-tree, or intersententially, if the full
sp-tree contains any {\sc period} operations. The third and fourth
sentences for Alt 13 in Figure~\ref{compare3-alt-fig} show the
conversion of a named restaurant ({\it Carmine's}) to a pronoun.

\begin{figure*}[htb]
  \centerline{\psfig{figure=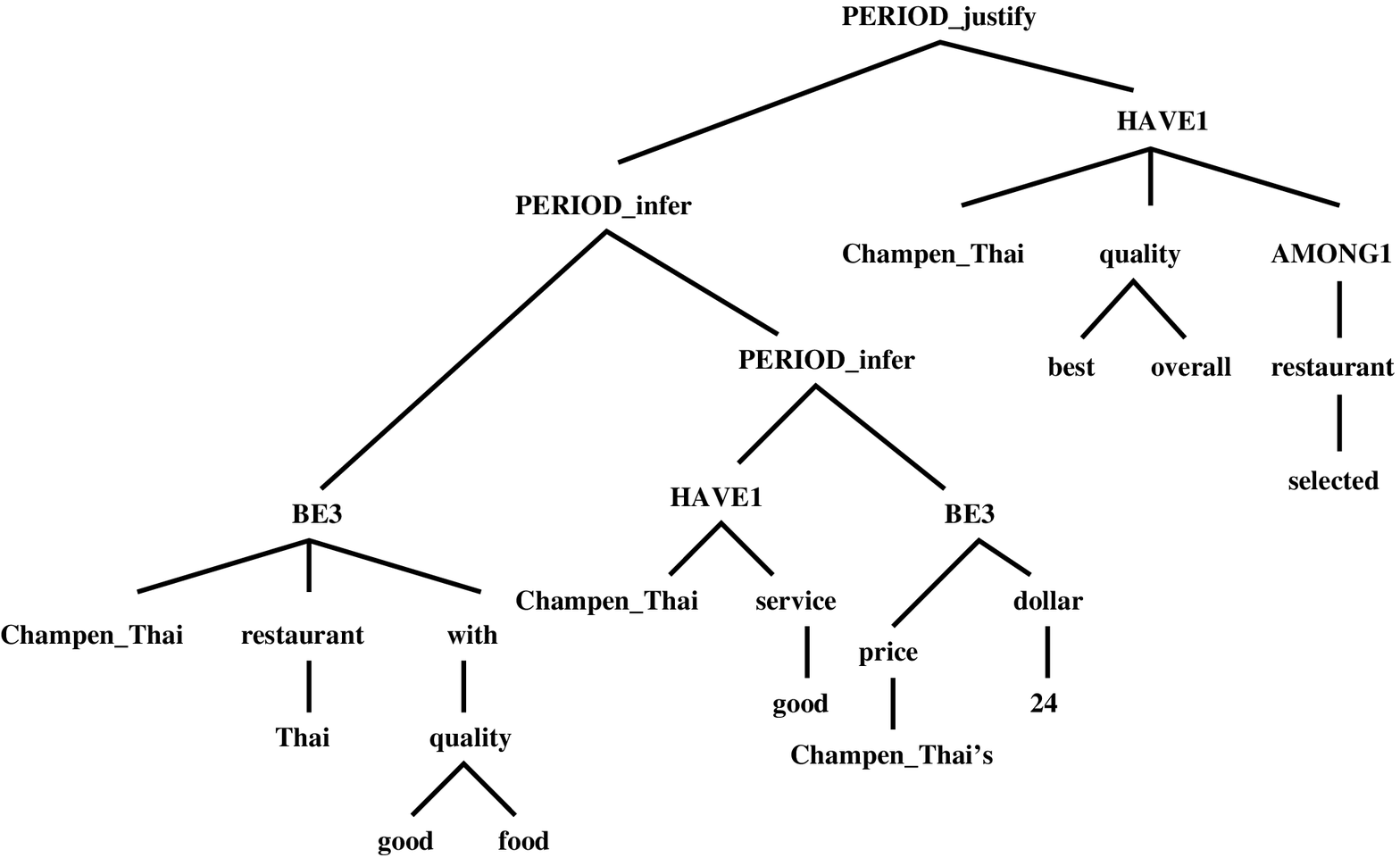,width=5.5in}}

\caption{Dependency tree for alternative 8 in
Figure~\ref{recommend-alt-fig}.} \label{alt8-dtree}
\end{figure*}

The {\bf sp-trees} for Alts 6 and 8 in Figure~\ref{recommend-alt-fig}
are shown in Figs.~\ref{alt6-sptree} and~\ref{alt8-sptree}.  Leaf
labels are concise names for assertions in the content plan, e.g.
{\bf assert-reco-best} is the claim (labelled 1) in
Figure~\ref{recommend-tplan-fig}. Because combination operations can
switch the order of their arguments, from satellite before nucleus
(SN) to nucleus before satellite (NS), the labels on the interior
nodes indicate whether this occurred, and specify the rhetorical
relation that the operation realizes.  These labels keep track of
the operations and substitutions used in constructing the tree and are
subsequently used in the tree feature set described in
Section~\ref{feat-gen-sec}, one of the feature sets
tested for training the {\spr}.
For example, the label at the root of the
tree in Figure~\ref{alt6-sptree} ({\bf CW-SINCE-NS-justify}) specifies
that the {\sc cw-conjunction} operation was used, with the {\it since}
cue word, with the nucleus first (NS), to realize the {\it justify}
relation.  Similarly, the bottom left-most interior node ({\bf
WITH-NS-infer}) indicates that the {\sc with-reduction} operation was
used, with the nucleus before the satellite (NS), to realize the {\it
infer} relation.

Figure~\ref{alt8-dtree} shows a d-tree for the content plan in
Figure~\ref{recommend-tplan-fig}.  This d-tree shows that the {\spg}
treats the {\sc period} operation as part of the lexico-structural
representation for the d-tree. The d-tree is split into multiple
d-trees at these nodes before being sent to RealPro for surface
realization.


Note that a tp-tree can have very different realizations, depending on
the operations of the {\spg}.  For example, the tp-tree in
Figure~\ref{alt6-tplantree} yields both Alt 6 and Alt 2 in
Figure~\ref{recommend-alt-fig}. Alt 2 is highly rated, with an average
human rating of 4.  However, Alt 6 is a poor realization of this plan,
with an average human rating of 2.5.

To summarize, {\sc SPaRKy}'s {\spg} transforms an input content plan into a
set of alternative pairs of sentence-plan trees and d-trees.  First,
assertions in the input content plan are grouped using principles from
centering theory.  Second, assertions are assigned to sentences and
discourse cues inserted using clause combining operations.  Third,
decisions about the realization of referring expressions are made on
the basis of recency.  The rhetorical relations and clause-combining
operations are domain-independent.

{\sc SPaRKy} uses two types of domain-dependent knowledge: the probability distribution over clause-combining operations, and the
d-trees that are input to the RealPro surface realizer.  In order to use {\sc SPaRKy} in a new domain, it might be necessary to:
\begin{itemize}
\item add new rhetorical relations if the content planner used additional rhetorical relations;
\item modify the probability distribution over clause-combining operations, either by hand or by learning from a corpus;
\item construct a new set of d-trees to capture the syntactic structure of sentences in the domain, unless we used a surface realizer that could take logical forms or semantic representations as input.
\end{itemize}

\section{Feature Generation}
\label{feat-gen-sec}

To train or use the {\spr}, each potential realization generated by
the SPG, along with its corresponding sp-tree and d-tree, is encoded
as a set of real-valued features (binary features are modeled with
values 0 and 1) from three
feature types:
\begin{itemize}
\item {\bf N-Gram features} -- simple word n-gram features generated from the realization of
{\spg} outputs;
\item {\bf Concept features} -- concept n-gram features generated from named entities in the realization of {\spg} outputs;
\item {\bf Tree features} -- these features represent
structural configurations in the sp-trees and d-trees output by the
SPG.
\end{itemize}
These features are automatically generated as described below.

\subsection{N-Gram Features}

N-gram features capture information about lexical selection and lexical
ordering in the realizations output by {\sc SPaRKy}.  A two-step approach is
used to generate these features.  First, a domain-specific rule-based
named-entity tagger (using {\sc MATCH}'s lexicons for restaurant, cuisine type
and location names) replaces specific tokens with their
types, e.g. {\it Babbo} with {\sc restname}.  Then, unigram, bigram
and trigram features and their counts are automatically generated.
The tokens {\it begin} and {\it end} indicate the beginning and end
of a realization.

N-gram feature names are prefixed with {\sc n-gram}.  For example, {\sc
ngram-cuisinename-restaurant-with} counts the occurrences of cuisine
type followed by ``restaurant'' and ``with'' (as in the realization
``Italian restaurant with''); {\sc ngram-begin-restname-which} counts
occurrences of realizations starting with a restaurant's name followed
by ``which''.  We also count words per presentation, and per sentence
in a presentation.

\subsection{Concept Features}

Concept features capture information about the concepts selected for a
presentation, and their linear order in the realization.  A two-step
approach is used to generate these features.  First, a named-entity
tagger marks the names of items in our restaurant database, e.g. {\it
Uguale}.  Then, unigram, bigram and trigram features and their counts
are automatically generated from the sequences of concepts in the
sentence plan for the realization.  As with the n-gram features,
the tokens {\it begin} and {\it end} indicate the beginning and end
of a realization.

Concept feature names are prefixed with {\sc conc}.  For example, {\sc
conc-decor-claim} is set to 1 if the claim is expressed directly after
information about decor, while the feature {\sc conc-begin-service}
characterizes utterances starting with information about service.  In
the concept n-gram features, we use '*' to separate individual
features.  We also count concepts per presentation, and
per sentence in a presentation.

\subsection{Tree Features}

Tree features capture declaratively the way in which {\sc merge,
infer} and {\sc cue-word} operations are applied to the tp-trees, and
were inspired by the parsing features used by \fullciteA{Collins00}.  They
count the occurrences of certain structural linguistic configurations
in the sp-trees and associated d-trees that the {\spg} generated.
Tree feature names are prefixed with {\sc r} for ``rule'' (sp-tree) or
{\sc s} for ``sentence'' (d-tree).

Several feature templates are used to generate tree features.  {\it
Local feature templates} record structural configurations local to a
particular node (its ancestors, daughters etc.); {\it global
feature templates}, used only for sp-tree features, record properties
of the entire sp-tree.

There are four types of local feature template: traversal features,
sister features, ancestor features and leaf features.  Traversal,
sister and ancestor features are generated for all nodes in sp-trees
and d-trees; leaf features are generated for sp-trees only.  The value
of each feature is the count of the described configuration in the
tree.  We discard features that occur fewer than 10 times to avoid
those specific to particular content plans.

For each node in the tree, {\bf traversal features} record the
preorder traversal of the subtree rooted at that node, for all
subtrees of all depths.  Feature names are the concatenation of the
prefix {\sc trav-}, with the names of the nodes (starting with the
current node) on the traversal path.  '*' is used to separate node
names.  An example is {\sc
r-trav-with-ns-infer*assert-reco-food-quality*assert-reco-cuisine}
(with value 1) of the bottom-left subtree in Figure~\ref{alt8-sptree}.

{\bf Sister features} record all consecutive sister nodes.  Names are
the concatenation of the prefix {\sc sis-}, with the names of the
sister nodes.  An example is {\sc
r-sis-assert-reco-best*cw-conjunction-infer} (with value 1) of the
tree in Figure~\ref{alt6-sptree}.

For each node in the tree, {\bf ancestor features} record all the
initial subpaths of the path from that node to the root.  Feature
names are the concatenation of the prefix {\sc anc-} with the
names of the nodes (starting with the current node).  An example is
{\sc r-anc-assert-reco-cuisine*with-ns-infer*cw-conjunction-infer}
(with value 1) of the tree in Figure~\ref{alt6-sptree}.

{\bf Leaf features} record all initial substrings of the frontier of
the sp-tree.  Names are the concatenation of the prefix {\sc leaf-},
with the names of the frontier nodes (starting with the current node).
For example, the sp-tree of Figure~\ref{alt6-sptree} has value 1 for
{\sc leaf-assert-reco-best} and also for {\sc
leaf-assert-reco-best*leaf-assert-reco-cuisine}, and the sp-tree of
Figure~\ref{alt8-sptree} has value 1 for {\sc
leaf-assert-reco-food-quality*assert-reco-cuisine}.

{\bf Global features} apply only to the sp-tree.  They record,
for each sp-tree and for each operation labeling a non-frontier
node, (1) the minimal number of leaves dominated by a node labeled
with that rule in that tree (MIN); (2) the maximal number of leaves
dominated by a node labeled with that rule (MAX); and (3) the average
number of leaves dominated by a node labeled with that rule (AVG).
For example, the sp-tree in Figure~\ref{alt6-sptree}
has value 4 for {\sc cw-conjunction-infer-max}, value 2 for {\sc cw-conjunction-infer-min}
and value 3 for {\sc cw-conjunction-infer-avg}.

\section{Training the Sentence Plan Ranker}
\label{spr-sec}

The {\spr} ranks alternative information presentations using a model
learned from user ratings of a set of training data.  The training
procedure is as follows:
\begin{itemize}

\item For each content plan in the training data, the {\spg} generates a
set of alternative sentence plans using a random selection of sentence
planning operators (Section~\ref{spg-sec});

\item Features are automatically generated from the surface
realizations and sentence plans so that each alternative sentence plan
is represented in terms of a number of real-valued features
(Section~\ref{feat-gen-sec});

\item Feedback as to the perceived quality of the realization of each
alternative sentence plan is collected from one or more users;

\item The RankBoost boosting method \fullcite{rankboost} learns a function
from the featural representation of each realization to its feedback,
that attempts to duplicate the rankings in the training examples.

\end{itemize}

We use RankBoost for three reasons.  First, it produces a ranking over
the input alternatives rather than a selection of one best
alternative.  Second, it can handle many sparse features.
Third, the function that it learns is a rule-based model showing the
effect of each feature on the ranking of the competing examples. These
models can be inspected and compared.  This allows us to qualitatively
analyze the models (Section~\ref{qual-results-sec}) in order to
understand the preferences of individuals, and the differences between
{\spr}s for individuals vs. groups.

This section describes the training of the {\spr} in detail.
The {\sc SPUR} content planner produces content plans for three dialogue
strategies:
\begin{itemize}
\item {\sc recommend}: recommend an entity from a set of entities
\item {\sc compare-2}: compare two entities
\item {\sc compare-3}: compare three or more entities
\end{itemize}
For each dialogue strategy, we start with a set of 30 representative
content plans from {\sc SPUR}.  The {\spg} was parameterized to produced up
to 20 distinct (sp-tree, d-tree) pairs for each content plan.  Each of
these was realized by RealPro.  Separately, we also obtained output
for each content plan from our template-based generator
(Section~\ref{template-sec}).

Both the {\sc SPaRKy} realizations and the template-based realizations were
randomly ordered and placed on a series of Web pages.  These 1830
realizations were then rated on a scale from 1 to 5 by the first two
authors of this paper, neither of whom had implemented the
template-based realizer or the {\spg}. The raters worked on this
rating task during sessions of one hour at a time for several hours a
day, over a period of a week. They were instructed to look at all 21
realizations for a particular content plan before rating any of them,
to try to use the whole rating scale, and to indicate their
spontaneous rating without repeatedly re-labelling the alternative
realizations.  They did not discuss their ratings or the basis for
their ratings at any time. Given the cognitive load and long duration
of this rating task, it was impossible for the raters to keep track of
which realizations came from {\sc SPaRKy} and which from the template-based
generator, and likely to be impossible to do more than generate a ``gestalt''
evaluation of each alternative.

Each (sp-tree, d-tree, realization) triple is an example input for
RankBoost; the ratings are used as feedback.  The experiments below
examine two uses of the ratings. First, we train and test an
{\spr} with the average of the ratings of the two users, i.e. we
consider the two users as representing a single user group. Second, we
train and test individualized {\spr}s, one for each user.

The {\spr} is trained using the RankBoost algorithm \fullcite{rankboost},
which we describe briefly here.  First, the training corpus is
converted into a set ${\cal T}$ of {\em ordered pairs} of examples
$x,y$:

\begin{center}
\parbox{0cm}{\begin{tabbing}
  ${\cal T} = \{ (x,y) |$ \=$x, y \mbox{ are alternatives for the same
    plan}$,\\
  \>$\mbox{$x$ is preferred to $y$ by user ratings}$\}
\end{tabbing}}
\end{center}

Each alternative realization $x$ is represented by a set of $m$
indicator functions $h_s(x)$ for $1 \leq s \leq m$. The indicator
functions are calculated by thresholding the feature values (counts)
described in Section~\ref{feat-gen-sec}.  For example, one indicator
function is:
 \[\small{
     h_{100}(x) = \left\{
         \begin{array}{ll}
         1 &  \mbox{if {\sc leaf-assert-reco-best}($x$) $\geq 1$}  \\
         0 &  \mbox{otherwise}
         \end{array}\right. ~ }
 \]

So $h_{100}(x)=1$ if the leftmost leaf is the assertion of the claim
as in Figure~\ref{alt6-sptree}. A single parameter $\alpha_s$ is
associated with each indicator function, and the ``ranking score''
for an example $x$ is calculated as
\[
F(x) = \sum_s \alpha_s h_s(x)
\]
This score is used to rank competing sp-trees of the same content plan
with the goal of duplicating the ranking found in the training data.
Training is the process of setting the parameters $\alpha_s$ to
minimize the following loss function:
\[
RankLoss = \frac{1}{|{\cal T}|} \sum_{(x,y) \in {\cal T}}eval(F(x) \leq
F(y))
\]
The {\it eval} function returns 1 if the ranking scores of the $(x,y)$
pair are misordered (so that $x$ is ranked higher than $y$ even though
in the training data $y$ is ranked higher than $x$), and 0
otherwise. In other words, the RankLoss is the percentage of
misordered pairs. As this loss function is minimized, the ranking
errors (cases where ranking scores disagree with human judgments) are
reduced. Initially all parameter values are set to zero. The
optimization method then greedily picks a single parameter at a time
-- the parameter which will make the most impact on the loss function
-- and updates the parameter value to minimize the loss.

In the experiments described below, we use two evaluation metrics:
\begin{itemize}
\item {\bf RankLoss}: The value of the training method's loss function;
\item {\bf TopRank}: The difference between the human rating of the top realization for each content plan and the human rating of the realization that the {\spr} predicts to be the top ranked.
\end{itemize}

\section{Quantitative Results}
\label{results-sec}

In this section, we describe three experiments with {\sc SPaRKy}:

\begin{enumerate}
\item {\bf Feature sets for trainable sentence planning:} We examine
which features (n-gram, concept, tree, all) lead to the best
performance for the sentence planning task, and find that n-gram
features sometimes perform as well as all the features.
\item {\bf Comparison with template-based generation:} We show that the
performance of a trainable sentence planner using the best performing
feature set is more consistent than that of a template-based
generator, although overall a template-based generator still performs
better.
\item {\bf Individualized sentence planners:} We show that people have
quite specific individual preferences regarding the three tasks of
sentence planning: information ordering, sentence aggregation, and use
of discourse cues; and furthermore, that a trainable sentence planner
can model these individual preferences. Moreover we show that in some
cases the individualized sentence planners are better than, or
statistically indistinguishable from, the template-based generator.

\end{enumerate}

We report results below separately for comparisons between two
entities and among three or more entities.  These two types of
comparison are generated using different strategies in the {\spg}, and
produce text that is very different both in terms of length and
structure.

\subsection{Feature Sets for Trainable Sentence Planning}
\label{cross-val-sec}

\begin{figure}[tb]
\begin{scriptsize}
\begin{center}
\begin{tabular}{|p{.1in}|p{3.0in}|c|c|c|c|c|} \hline
Alt & Realization & A & B & SPR$_A$ & SPR$_B$ & SPR$_{AVG}$\\ \hline 
6 &  Chanpen Thai has the best overall quality among the selected
restaurants since it is a Thai restaurant, with good service, its
price is 24 dollars, and it has good food quality. & 1 & 4  & 0.16 & 0.65 & 0.58\\ \hline %
7 & Chanpen Thai has the best overall quality among the selected restaurants because it has good service, it has good food quality, it is a Thai restaurant, and its price is 24 dollars. & 2 & 5 & 0.38 & 0.54 & 0.42\\ \hline %

4 & Chanpen Thai has the best overall quality among the selected
restaurants. It has good food quality, with good service, it is a
Thai restaurant, and its price is 24 dollars. & 2 & 4 & 0.53 & 0.62
& 0.53\\ \hline

9 & Chanpen Thai is a Thai restaurant, with good food
quality, its price is 24 dollars, and it has good service. It has
the best overall quality among the selected restaurants. & 2 & 4 &
0.47 & 0.53 & 0.63\\ \hline

5 & Chanpen Thai has the best overall quality
among the selected restaurants. It has good service. It has good
food quality. Its price is 24 dollars, and it is a Thai restaurant.
& 3 & 2 & 0.59 & 0.32 & 0.46\\ \hline

3 & Chanpen Thai has the best
overall quality among the selected restaurants. Its price is 24
dollars. It is a Thai restaurant, with good service. It has good
food quality. & 3 & 3 & 0.64 & 0.40 & 0.62\\ \hline

10 & Chanpen Thai has
the best overall quality among the selected restaurants. It has good
food quality. Its price is 24 dollars. It is a Thai restaurant, with
good service. & 3 & 3 & 0.67 & 0.46 & 0.58 \\ \hline

2 & Chanpen Thai has
the best overall quality among the selected restaurants. Its price
is 24 dollars, and it is a Thai restaurant. It has good food quality
and good service. & 4 & 4 & 0.75 & 0.50 & 0.74\\ \hline

1 & Chanpen Thai
has the best overall quality among the selected restaurants. This
Thai restaurant has good food quality. Its price is
24 dollars, and it has good service. & 4 & 3  & 0.64 & 0.52 & 0.45\\ \hline %

8 & Chanpen Thai is a Thai restaurant, with good food quality. It
has good service. Its price is 24 dollars. It has the best overall
quality
among the selected restaurants. & 4 & 2    & 0.81 & 0.29 & 0.73\\ \hline %
\end{tabular}
\end{center}
\end{scriptsize}
\caption{ Some alternative realizations for the content plan in
Figure~\ref{recommend-tplan-fig}, with feedback from users A and B
(1$=$worst and 5$=$best) and rankings from the trained SPRs for
users A and B and mean(A,B) ($[0,1]$).}
\label{recommend-alt-spr-fig}
\end{figure}

Using a cross-validation methodology, we repeatedly train the {\spr}
on a random 90\% of the corpus, and test on the remaining
10\%.  Here, we use the averaged feedback from user A and user B as
feedback.  Figure~\ref{recommend-alt-spr-fig} repeats the examples in
Figure \ref{recommend-alt-fig}, here showing both the user rankings
and the rankings for a ranking function that was learned by the
trained {\spr}s for both users A and B and for the AVG user.

Table \ref{ranking-loss-feats} shows RankLoss for each feature
set (Section \ref{feat-gen-sec}). Paired
t-tests comparing the ranking loss for different feature sets show
surprisingly few performance differences among the features. Using all
the features ({\bf All}) always produces the best results, but the
differences are not always significant.

The n-gram features give results comparable to all the features for
both {\sc compare-2} and {\sc recommend}.  An analysis of the learned
models suggests that one reason that n-gram features perform well is
because there are individual lexical items that are uniquely
associated with many of the combination operators, such as the lexical
item {\it with} for the {\sc with-ns} operator.  This means that
the detailed representations of the content and structure of an
information presentation as represented by the tree features
are equivalent to n-gram features in this application domain.

The concept features always perform worse than all the features,
indicating that the linear ordering of concepts only accounts for some
of the variation in rating feedback. For the two types of comparison,
performance using the concept features approaches that of the other
feature sets. However, for recommendations, performance using the
concept features is much worse than that using n-gram features or all
the features.  In the qualitative analysis presented in Section
\ref{qual-results-sec}, we discuss some aspects of the models for
recommendations that might account for this large difference in
performance.

\begin{table}[htb]
\begin{center}

\begin{tabular}{|l|c|c|c|}
\hline
{\bf Feature set/Strategy}                   &  {\bf {\sc compare-2}}    &       {\bf {\sc compare-3}} &  {\bf
      {\sc recommend}}\\
\hline
{\bf Random Baseline}        & 0.50 & 0.50 & 0.50 \\
\hline
{\bf Concept}        & 0.16 ($p < .000$)& 0.16 ($p < .021$)& 0.32 ($p < .000$)\\
\hline

{\bf N-Gram}        & {\bf 0.14} ($p < .161$) & 0.15 ($p < .035$)& {\bf 0.21} ($p < .197$)\\
\hline
{\bf Tree }   & 0.14 ($p < .087$) & 0.16 ($p < .007$)& 0.22 ($p < .001$)\\
\hline
{\bf All}         & {\bf 0.13} & {\bf 0.14} & {\bf 0.20} \\
\hline
\end{tabular}
\end{center}
\caption{AVG model's ranking error with different feature sets, for
all strategies. Results are averaged over 10-fold cross-validation,
testing over the mean feedback. $p$ values in parentheses indicate the
level of significance of the decrease in accuracy when compared to the
model using all the features.  Cases where different feature sets
perform as well as all the features are marked in bold.
\label{ranking-loss-feats}}
\end{table}

Table \ref{toprank-results-fig} shows results with all the features
using the TopRank evaluation metric, calculated for two-fold
cross-validation, to be comparable with previous work
\fullcite{WRR02,Stentetal04}.\footnote{The TopRank metric is sensitive to
the distribution of ranking feedback and {\spr} scores in the test
set, which means that it is sensitive to the number of
cross-validation folds.}  We evaluated {\sc SPaRKy} on the test sets by
comparing three data points for each content plan: Human (the score of
the best sentence plan that {\sc SPaRKy}'s {\spg} can produce as selected by the
human users); {\sc SPaRKy} (the score of the {\spr}'s top-ranked selected
sentence); and Random (the score of a sentence plan randomly selected
from the alternative sentence plans).  For all three presentation
types, a paired t-test comparing {\sc SPaRKy} to Human to Random showed that
{\sc SPaRKy} was significantly better than Random ($df = 59$, $p < .001$)
and significantly worse than Human ($df = 59$, $p < .001$). The
difference between the {\sc SPaRKy} scores and the Human scores indicates
how much performance could be improved if the {\spr} were perfect at
replicating the Human ratings.


\begin{table}[htb]
\begin{center}
\begin{tabular}{|llccc|} \hline
User & Strategy & {\sc SPaRKy}  & Human & Random   \\ \hline \hline
AVG & {\sc recommend} & 3.6 (0.77) & 3.9 (0.55) & 2.8 (0.81) \\ \hline %
AVG & {\sc compare-2} & 4.0 (0.66) & 4.4 (0.54) &  2.8 (1.30)  \\ \hline %
AVG & {\sc compare-3} & 3.6 (0.68) & 4.0 (0.49) &  2.7 (1.20)  \\ \hline %
\end{tabular}
\end{center}
\caption{TopRank scores for {\sc recommend}, {\sc compare-2} and {\sc compare-3} (N =
 180), using all the features, for {\sc SPaRKy} trained on AVG feedback, with standard deviations.
\label{toprank-results-fig}}
\end{table}

\subsection{Comparison with Template Generation}
\label{template-results-sec}

\begin{table}[htb]
\begin{center}
\begin{tabular}{|llccc|} \hline
User & Strategy & {\sc SPaRKy}  & Human & Template   \\ \hline \hline
AVG & {\sc recommend} & 3.6 (0.59) & 4.4 (0.37) & 4.2 (0.74) \\ \hline %
AVG & {\sc compare-2} & 3.9 (0.52) & 4.6 (0.39) &  3.6 (0.75)  \\ \hline %
AVG & {\sc compare-3} & 3.4 (0.38) & 4.6 (0.35) &  4.1 (1.23)  \\ \hline %
\end{tabular}
\end{center}
\caption{ TopRank scores for {\sc MATCH}'s template-based generator, {\sc SPaRKy}(AVG) and Human. N = 180, with standard deviations. \label{template-results-fig}}
\end{table}



As described above, the raters also rated the single output of the
template-based generator for {\sc MATCH} for each content plan in the
training data.  Table~\ref{template-results-fig} shows the mean
TopRank scores for the template-based generator's output (Template),
compared to the best plan the trained {\spr} selects ({\sc SPaRKy}), and the
best plan as selected by a human oracle (Human). In each fold, both
{\sc SPaRKy} and the Human oracle select the best of 10 sentence plans for
each text plan, while the template-based generator produces a single
output with a single human-rated score. A paired t-test comparing
Human with Template shows that there are no significant differences
between them for {\sc recommend} or {\sc compare-3}, but that Human is
significantly better for {\sc compare-2} ($df = 29$, $t = 4.8$, $p <
.001$).  The users evidently did not like the {\sc compare-2} template.
A paired t-test comparing {\sc SPaRKy} to Template
shows that the template-based generator is significantly better
for both {\sc recommend} and {\sc compare-3} ($df = 29$, $t
= 2.1$, $p < .05$), while there is a trend for {\sc SPaRKy} to be better for
{\sc compare-2} ($df = 29$, $t = 2.0$, $p = .055$).

Also, the standard deviation for Template strategies is
wider than for Human or {\sc SPaRKy}, indicating that while the
template-based generator performs well overall, it performs poorly on
some inputs. One reason for this might be that {\sc SPUR}'s
decision-theoretic user model selects a wide range and number of
content items for different users, and for conciseness settings
(See Figures~\ref{concise-recommends-fig} and
~\ref{concise-compares-fig}). This means that it is difficult to
handcraft a template-based generator to handle all the different cases
well.

The gap between the Human scores (produced by the {\spg} but selected
by a human rather than by the {\spr}) and the Template scores shows
that the {\spg} produces sentence plans as good as those of the
template-based generator, but the accuracy of the {\spr} needs to be
improved. Below, Section~\ref{individual-results-sec} shows that when
the {\spr} is trained for individuals, {\sc SPaRKy}'s performance is
indistinguishable from the template-based generator in most cases.

\subsection{Comparing Individualized Models to Group Models}
\label{individual-results-sec}

We discussed in Section \ref{introsec} that the differences in the
rating feedback from users A and B for competing realizations (See
Figure~\ref{recommend-alt-fig}) suggest that each user
has different perceptions as to the quality of the potential
realizations.  To quantify the utility and the feasibility of training
individualized {\spr}s, we first examine the feasibility of training
models for individual users.

 \begin{figure}[h]\begin{center}\includegraphics[angle=-90,
 width=28pc]{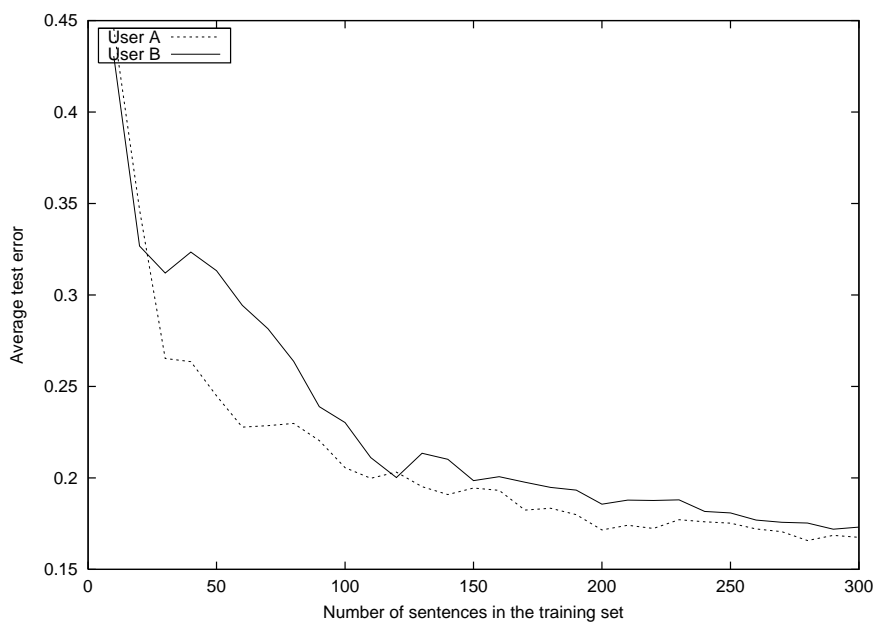}
\vspace{-0.10in} \caption{Variation of the
 testing error for both users as a function of the number of training
utterances.}\label{plot_feedback}
 \end{center}\end{figure}

The results in Table~\ref{ranking-loss-feats} are
based on a corpus of 600 examples, rated by each user, which may
involve too much effort for most users.
We would like to know whether a high-performing individualized {\spr}
can be trained from less labelled data.  Figure~\ref{plot_feedback}
plots ranking error rates as a function of the amount of training
data.
This data suggests that error rates around 0.20 could be
acquired with a much smaller training set, i.e. with a training set of
around 120 examples, which is certainly more feasible.

\begin{table}[htb]
\begin{center}
\begin{tabular}{|c|c|c|c|}
\hline
{\sc recommend}  &   A's model   &   B's model   &   AVG model\\
\hline \hline
A's test data   &   0.17  &   0.52  &   0.29    \\
\hline
B's test data   &   0.52   &   0.17  &   0.27    \\
\hline
AVG's test data &   0.31   &   0.31  &   0.20    \\
\hline
\end{tabular}
\end{center}
\caption{Ranking error for various configurations with the {\sc recommend} strategy.
\label{indiv-rec-table}}
\end{table}

\begin{table}[htb]
\begin{center}
\begin{tabular}{|c|c|c|c|}
\hline
{\sc compare-2}  &   A's model   &   B's model   &   AVG model\\
\hline \hline
A's test data   &   0.16  &   0.26 &   0.20    \\
\hline
B's test data   &   0.23  &   0.11  &   0.13    \\
\hline
AVG's test data &   0.17  &   0.16  &   0.13    \\
\hline
\end{tabular}
\end{center}
\caption{Ranking error for various configurations with the {\sc compare-2} strategy.
\label{indiv-comp2-table}}
\end{table}

\begin{table}[htb]

\begin{center}
\begin{tabular}{|c|c|c|c|}
\hline
{\sc compare-3}  &   A's model   &   B's model   &   AVG model\\
\hline \hline
A's test data   &   0.13  &   0.30  &   0.18    \\
\hline
B's test data   &   0.26  &   0.14  &   0.18    \\
\hline
AVG's test data &   0.17  &   0.20  &   0.14    \\
\hline
\end{tabular}
\end{center}
\caption{Ranking error for various configurations with the {\sc compare-3} strategy.
\label{indiv-comp3-table}}
\end{table}




We then examine if trained individualized {\spr}s are accurate.  The
results in Tables \ref{indiv-rec-table}, \ref{indiv-comp2-table} and
\ref{indiv-comp3-table} show RankLoss for several training and testing
configurations for each strategy (using 10-fold cross-validation). We
compare the two individualized models with models trained on A and B's
mean feedback (AVG).  For each model, we test on its own test data,
and on test data for the other models. This shows how
well a model might `fit' if customizing an {\spr} to a new domain or
user group.  For example, if we train a model for recommendations
using feedback from a group of users, and then deploy this system to
an individual user, we might expect model fit differences similar to
those in Table~\ref{indiv-rec-table}.

Of course, there may be strongly conflicting preferences in any group
of users.  For example, consider the differences in the ratings for
users A and B and the average ratings in
Figure~\ref{recommend-alt-fig}. Alt-1 and Alt-7 are equivalent using
the average feedback, but user A dislikes Alt-7 and likes Alt-1 and
vice versa for user B. Column 3 of Table~\ref{indiv-rec-table} shows
that the average model, when used in an {\spr} for user A or user B
has a much higher ranking error (.29 and .27 respectively) than that
of an {\spr} customized to user A (.17 error) or customized to user B
(.17 error).

An examination of Tables \ref{indiv-rec-table},
\ref{indiv-comp2-table} and \ref{indiv-comp3-table} shows that in
general, there are striking differences between models trained and
tested on one individual's feedback (RankLoss ranges from 0.11 to
0.17) and cross-tested models (RankLoss ranges from 0.13 to 0.52).
Also, the average (AVG) models always perform more poorly for both
users A and B than individually-tailored models. As a baseline for
comparison, a model ranking sentence alternatives randomly produces an
error rate of 0.5 on average; Table \ref{indiv-rec-table} shows that
models trained on one user's data and tested on the other's can
perform as badly as the random model baseline. This suggests that
the differences in the users' ratings are not random noise.


In some cases, the average model also performs significantly worse
than the individual models even when tested on feedback from the
``average'' user (the diagonal in Tables
\ref{indiv-rec-table}, \ref{indiv-comp2-table} and \ref{indiv-comp3-table}).  This
suggests that in some cases it is harder to get a good model for the
average user case, possibly because the feedback is more inconsistent.
For recommendations, the performance of each individual model
is significantly better than the average model ($df = 9$,
$t = 2.6$, $p < .02$). For {\sc compare-2} the average model is better
than user A's ($df = 9$, $t = 2.3$, $p < .05$), but user B's model is
better than the average model ($df=9$, $t = 3.1$, $p < .01$).

\begin{table}[htb]
\begin{center}
\begin{tabular}{|llccc|} \hline
User & Strategy & {\sc SPaRKy}  & Human & Template   \\ \hline \hline
A & {\sc recommend} & 3.5 (0.87) & 3.9 (0.61) & 3.9 (1.05) \\ \hline %
A & {\sc compare-2} & 3.8 (0.98) & 4.3 (0.73) &  4.2 (0.64)  \\ \hline %
A & {\sc compare-3} & 3.1 (1.02) & 3.6 (0.80) &  3.9 (1.19)  \\ \hline  \hline %
B & {\sc recommend} & 4.4 (0.70) & 4.7 (0.46) & 4.5 (0.76) \\ \hline %
B & {\sc compare-2} & 4.4 (0.69) & 4.7 (0.53) &  3.1 (1.21)  \\ \hline %
B & {\sc compare-3} & 4.4 (0.62) & 4.8 (0.40) &  4.2 (1.34)  \\ \hline %
\end{tabular}
\caption{TopRank scores for the Individualized {\sc SPaRKy} as compared
with {\sc MATCH}'s
template-based generator as rated separately by Users A and B, and individual User A and User B Human Oracles. Standard Deviations are in parentheses.
 N =
180.} \label{indiv-template-results-fig}
\end{center}
\end{table}

We can also compare the template-based generator to the individualized
{\sc SPaRKy} generators using the TopRank metric (See Table
\ref{indiv-template-results-fig}). All comparisons are done with
paired t-tests using the Bonferroni adjustment for multiple
comparisons.

For {\sc recommend}, there are no significant differences between
{\sc SPaRKy} and Template for User A ($df=59$, $t = 2.3$, $p = .07$), or for
User B ($df=59$, $t = 1.6$, $p = .3$). There are also no significant
differences for either user between Template and Human ($df=59$, $t <
1.5$, $p > 0.4$).

For {\sc compare-2}, there are large differences between Users A and B.
User A appears to like the template for {\sc compare-2} (average rating
is 4.2) while User B does not (average rating is 3.1).  For User A,
there are no significant differences between {\sc SPaRKy} and Template
($df=59$, $t = 2.3$, $p = .07$), and between Template and Human
($df=59$, $t = 0.1$, $p = .09$), but User B strongly prefers {\sc SPaRKy} to
Template ($df=59$, $t = 7.7$, $p < .001$).

For {\sc compare-3}, there are also large differences between Users A
and B.  User A likes the template for {\sc compare-3} (average rating
3.9), and strongly prefers it to the individualized {\sc SPaRKy} (average
rating 3.1) ($df=59$, $t = 3.4$, $p < .004$). User B also likes the
template (average rating 4.2), but there are no significant
differences with {\sc SPaRKy} (average rating 4.4) ($df=59$, $t = 1.0$, $p =
.95$).

For both users, and for every strategy, even with individually trained
{\spr}s, there is still a significant gap between {\sc SPaRKy} and Human
scores, indicating that the performance of the {\spr} could be
improved ($df=59$, $t = 3.0$, $p < .006$).

These results demonstrate that {\it trainable sentence
planning can produce output comparable to or better than that of a
template-based generator}, with less programming effort and more
flexibility.

\section{Qualitative Analysis}
\label{qual-results-sec}


An important aspect of RankBoost is that the learned models are
expressed as rules: a qualitative examination of the learned models
may highlight individual differences in linguistic preferences, and
help us understand why {\sc SPaRKy}'s {\spg} can produce sentence plans that
are better than those produced by the template-based generator, and
why the individually trained {\spr}s usually select sentence plans
that are as good as the templates.  To qualitatively compare the
learned ranking models for the individualized {\spr}s, we assess both
which linguistic aspects of an utterance (which features) are
important to an individual, and how important they are.  We evaluate
whether an individual is oriented towards a particular feature by
examining which features' indicator functions $h_s(x)$ have non-zero
values.  We evaluate how important a feature is to an individual by
examining the magnitude of the parameters $\alpha_s$.

There are two potential problems with this approach, The first problem
is that the feature templates produce thousands of features, some of
which are redundant, so that differences in each model's indicator
functions can be spurious.  Therefore, to allow more meaningful
qualitative comparisons between models, one of a pair of perfectly
correlated features is eliminated.

The second problem arises from RankBoost's greedy algorithm. The
selection of which parameter $\alpha_s$ to set on any round of
boosting is highly dependent on the training set, so that the models
derived from a single episode of training are highly variable.  To
compare indicator functions independently of the training set, we
adopt a bootstrapping method to identify a feature set for each user
that is independent of a particular training episode. By repeatedly
randomly selecting 10 alternatives for training and 10 for testing for
each content plan, we created 50 different training sets for each
user.  We then average the $\alpha$ values of the features selected by
RankBoost over these 50 training runs, and conduct experiments using
only the 100 features for each user with the highest average $\alpha$
magnitude.  In Section \ref{bootstrap-feat-sec} we discuss differences
in the types of feature that are selected by the bootsrapping
algorithm just outlined.  Section \ref{indiv-sec} discusses
differences in models produced using the tree features for user A and
user B, while section \ref{average-sec} discusses differences between
the average model and the individual models.


\subsection{Types of Bootstrapped Features}
\label{bootstrap-feat-sec}

The bootstrapping process selects a total of 100 features for each
strategy and for each type of feedback (individual or averaged).  We
found differences in the features along both dimensions.

\begin{table}
\centering
\begin{tabular}{|l|l|c|c|c|c|c|} \hline
Model & Strategy & \multicolumn{5}{|c|}{Feature Type} \\ \hline
 & & Tree & N-Gram & Concept & Leaf & Global \\ \hline
AVG & {\sc recommend} & 45 & 36 & 9 & 7 & 3 \\
 & {\sc compare-2} & 37 & 46 & 12 & 1 & 4 \\
 & {\sc compare-3} & 63 & 29 & 4 & 1 & 3 \\ \hline
A & {\sc recommend} & 50 & 29 & 14 & 4 & 3 \\
 & {\sc compare-2} & 35 & 51 & 10 & 3 & 1 \\
 & {\sc compare-3} & 47 & 37 & 11 & 1 & 4 \\ \hline
B & {\sc recommend} & 47 & 34 & 9 & 6 & 4 \\
 & {\sc compare-2} & 45 & 36 & 13 & 1 & 5 \\
 & {\sc compare-3} & 47 & 34 & 9 & 6 & 4 \\ \hline
\end{tabular}
\caption{\label{featcounts}Features in the top 100 with the highest
average $\alpha$ for each user model.}
\end{table}

Table~\ref{featcounts} shows the number of features of each type that
were in the top 100 (averaged over 50 training runs).  Only 9 features
are shared by the three strategies for the AVG model; these shared
features are usually n-gram features.  For User A, 6 features are
shared by the three strategies (mostly n-gram features).  For User B,
there are no features shared by the three strategies.

We also found that some features capture specific interactions between
domain-specific content items and syntactic structure, which are
difficult to model in a rule-based or template-based generator. An
example is Rule (1) in Figure \ref{amanda-rule-fig} which
significantly lowers the ranking of any sentence plan in which
neighborhood information ({\sc assert-reco-nbhd}) is combined with
subsequent content items via the {\sc with-ns} operation.  Among the
bootstrapped features for the average user, 16 features for {\sc
compare-2} count interactions between domain-specific content and
syntactic structure. For {\sc compare-3}, 22 features count such
interactions, and the bootstrapped features for {\sc recommend}
include 39 such features.  We examine some of the models derived from
these features in detail below.

\subsection{Differences in Individual Models}
\label{indiv-sec}



\begin{figure}[htb]
\begin{center}
\begin{small}\begin{tabular}{|c p{5.0in} c|}
\hline
N & Condition & $\alpha$ \\
\hline
1  & {\sc r-anc-assert-reco-nbhd*with-ns-infer} $\geq 1$ & -1.26 \\ \hline
2  & {\sc cw-conjunction-infer-avg-leaves-under} $\geq 3.1$ & -0.58 \\ \hline
3  & {\sc r-anc-assert-reco*with-ns-infer*cw-conjunction-infer} $\geq 1$ & -0.33 \\ \hline
4  & {\sc leaf-assert-reco-best*assert-reco-price} $\geq 1$ & -0.29 \\ \hline
5  & {\sc cw-conjunction-infer-avg-leaves-under} $\geq 2.8$ & -0.27 \\ \hline
6  & {\sc r-trav-with-ns-infer*assert*assert} $\geq 1$ & -0.22 \\ \hline

7  & {\sc r-anc-cw-conjunction-infer*cw-conjunction-infer} $\geq 1$ & -0.17 \\ \hline
8  & {\sc with-ns-infer-min-leaves-under} $\geq 1$ & -0.13 \\ \hline
9  & {\sc r-anc-assert-reco*with-ns-infer} $\geq 1$ & -0.11 \\ \hline
10 & {\sc cw-conjunction-infer-max-leaves-under} $\geq 3.5$ & -0.07 \\ \hline
11 & {\sc r-trav-with-ns-infer*assert-reco*assert-reco} $\geq 1$ & -0.07 \\ \hline
12 & {\sc r-anc-assert*with-ns-infer} $\geq 1$ & -0.03 \\ \hline
13 & {\sc r-anc-with-ns-infer*relative-clause-infer} $\geq 1$ & -0.01 \\ \hline
14 & {\sc r-anc-assert*with-ns-infer*relative-clause-infer} $\geq 1$ & -0.01 \\ \hline
15 & {\sc cw-conjunction-infer-avg-leaves-under} $\geq 4.1$ & -0.01 \\ \hline

16 & {\sc r-anc-assert-reco-cuisine*with-ns-infer*period-infer} $\geq 1$ & 0.10 \\ \hline

17 & {\sc cw-conjunction-infer-avg-leaves-under} $\geq 2.2$ & 0.15 \\ \hline
18 & {\sc r-anc-assert-reco-food-quality*merge-infer} $\geq 1$ & 0.18 \\ \hline
19 & {\sc r-anc-assert-reco*merge-infer} $\geq 2.5$ & 0.20 \\ \hline
20 & {\sc r-anc-assert-reco-decor*merge-infer} $\geq 1$ & 0.22 \\ \hline
21 & {\sc r-anc-assert*merge-infer} $\geq 2.5$ & 0.25 \\ \hline
22 & {\sc r-trav-merge-infer} $\geq 1.5$ & 0.27 \\ \hline
23 & {\sc r-trav-with-ns-infer*assert-reco-service*assert-reco-food-quality} $\geq 1$ & 0.40 \\ \hline
24 & {\sc leaf-assert-reco-food-quality*assert-reco-cuisine} $\geq 1$ & 0.46 \\ \hline
25 & {\sc cw-conjunction-infer-avg-leaves-under} $\geq 3.8$ & 0.46 \\ \hline
26 & {\sc leaf-assert-reco-food-quality} $\geq 1$ & 0.60 \\ \hline
27 & {\sc s-trav-have1*propernoun-restaurant*II-quality*attr-among1} $\geq 1$ & 0.68 \\ \hline
28  & {\sc s-anc-attr-with*have1} $\geq 1$ & 0.71 \\ \hline
\end{tabular}\end{small}
\end{center}
\caption{\label{amanda-rule-fig} A subset of rules and corresponding $\alpha$
  values of User A's model, ordered by $\alpha$.}\end{figure}

\begin{figure}[htb]
\begin{center}
\begin{small}\begin{tabular}{|c p{5.1in} c|}
\hline
N & Condition & $\alpha$ \\
\hline
1  & {\sc r-sis-assert-reco-relative-clause-infer} $\geq 1$ & -1.01 \\ \hline
2  & {\sc r-sis-period-infer-assert-reco} $\geq 1$ & -0.71 \\ \hline

3  & {\sc r-anc-assert-reco-nbhd*with-ns-infer} $\geq 1$ & -0.50 \\ \hline
4  & {\sc r-anc-assert-reco*period-infer*period-infer} $\geq 1.5$ & -0.49 \\ \hline
5  & {\sc r-anc-assert-reco-food-quality*with-ns-infer*relative-clause-infer} $\geq 1$ & -0.41 \\ \hline
6  & {\sc r-anc-assert-reco-cuisine*with-ns-infer*relative-clause-infer} $\geq 1$ & -0.39 \\ \hline
7  & {\sc r-anc-assert-reco*period-infer} $\geq 1$ & -0.35 \\ \hline
8  & {\sc leaf-assert-reco-price} $\geq 1$ & -0.32 \\ \hline
9 & {\sc r-anc-assert*period-infer*period-infer} $\geq 1.5$ & -0.26 \\ \hline

10  & {\sc leaf-assert-reco-decor} $\geq 1$ & -0.14 \\ \hline
11  & {\sc r-anc-assert*relative-clause-infer*period-infer} $\geq 1.5$ & -0.07 \\ \hline
12  & {\sc r-trav-relative-clause-infer*assert-reco*with-ns-infer} $\geq 1$ & -0.05 \\ \hline
13  & {\sc cw-conjunction-infer-avg-leaves-under} $\geq 3.1$ & -0.03 \\ \hline
14  & {\sc cw-conjunction-infer-avg-leaves-under} $\geq 3.3$ & -0.03 \\ \hline
15  & {\sc cw-conjunction-infer-avg-leaves-under} $\geq 2.2$ & -0.01 \\ \hline
16  & {\sc r-anc-assert*relative-clause-infer*period-infer} $\geq 1$ & 0.03 \\ \hline
17  & {\sc leaf-assert-reco-service} $\geq 1$ & 0.07 \\ \hline
18  & {\sc s-trav-attr-with} $\geq 1$ & 0.18 \\ \hline
19  & {\sc r-anc-assert-reco-cuisine*with-ns-infer*cw-conjunction-infer} $\geq 1$ & 0.27 \\ \hline
20  & {\sc cw-conjunction-infer-avg-leaves-under} $\geq 3.6$ & 0.36 \\ \hline
21  & {\sc leaf-assert-reco-best} $\geq 1$ & 0.47 \\ \hline
22  & {\sc leaf-assert-reco-best*assert-reco-cuisine} $\geq 1$ & 0.50 \\ \hline
23  & {\sc cw-conjunction-infer-avg-leaves-under} $\geq 2.8$ & 0.52 \\ \hline
24  & {\sc r-trav-with-ns-infer*assert-reco-cuisine*assert-reco-food-quality} $\geq 1$ & 0.76 \\ \hline

\end{tabular}\end{small}
\end{center}
\caption{\label{lyn-rule-fig}A subset of rules and corresponding $\alpha$
  values of User B's model, ordered by $\alpha$.}\end{figure}

To further analyze individual linguistic preferences for information
presentation strategies, we now qualitatively compare the two models
for Users A and B.  We believe that this qualitative analysis provides
additional evidence that the differences in the users' ranking
preferences are not random noise. We identify differences among the
features selected by RankBoost, and their $\alpha$ values, using
models derived using bootstrapping over the tree features only, since
they are easier to interpret qualitatively. Of course many different
models are possible.  User A's model consists of 109 rules; a subset
are in Figure~\ref{amanda-rule-fig}.  User B's model consists of
90 rules, a subset of which are shown in Figure~\ref{lyn-rule-fig}.  We first
consider how the individual models account for the rating differences
for Alt-6 and Alt-8 from Figure~\ref{recommend-alt-fig} (repeated in
Figure~\ref{recommend-alt-spr-fig} with ratings from the trained
{\spr}s), and then discuss other differences.

{\bf Comparing Alt-6 and Alt-8:}
Alt-6 is highly ranked by User B but not by User A. Alt-6
instantiates Rule 21 
of Figure~\ref{lyn-rule-fig}, expressing User B's preferences about
linear order of the content. (Alt-6's sp-tree is in
Figure~\ref{alt6-sptree}.)  Rule 21 increases the rating of examples
in which the claim, i.e. {\sc assert-reco-best} ({\it Chanpen Thai has
the best overall quality}), is realized first. Thus, unlike user A,
user B prefers the claim at the beginning of the utterance (the
ordering of the claim is left unspecified by argumentation theory
\fullcite{careninimoore00b}).  Rule 22 increases the rating of examples in
which the initial claim is immediately followed by the type of cuisine
({\sc assert-reco-cuisine}).  These rules interact with Rule 19 in
Figure~\ref{lyn-rule-fig}, which specifies a preference for
information following {\sc assert-reco-cuisine} to be combined via the
{\sc with-ns} operation, and then conjoined ({\sc
cw-conjunction-infer}) with additional evidence.  Alt-6 also
instantiates Rule 23 in User B's model, with an $\alpha$ value of .52
associated with multiple uses of the {\sc cw-conjunction-infer}
operation.


User A's low rating of Alt-6 arises
from A's dislike of the {\sc with-ns} operation (Rules 3, 8, 9, 11 and
12) and the {\sc cw-conjunction-infer} operation (Rules 3, 5, 7, 10
and 15) in Figure~\ref{amanda-rule-fig}. (Contrast User B's Rule 23 with
User A's Rules 5 and 17.) Alt-6 also
fails to instantiate A's preference for food quality and cuisine
information to occur first (Rules 24 and 26).
Finally, user A also prefers the claim {\sc assert-reco-best} to be realized
in its own sentence (Rule 27).

By contrast, {\bf Alt-8} is rated highly by User A
but not by User B (see Figure~\ref{recommend-alt-fig}).
Even though Alt-8 instantiates the negatively evaluated {\sc with-ns}
operation (Rules 3, 8, 9 and 11 in Figure~\ref{amanda-rule-fig}), there are no
instances of {\sc cw-conjunction-infer} (Rules 3, 5, 7, 10 and 15).
Moreover Alt-8 follows A's ordering preferences (Rules 24 and 26)
which describe sp-trees with {\sc assert-reco-food-quality} on the
left frontier, and trees where it is followed by {\sc
assert-reco-cuisine}. (See Alt-8's sp-tree in
Figure~\ref{alt8-sptree}.) 
Rule 27 also increases the rating of Alt-8 with its large positive
$\alpha$ reflecting the expression of the claim in its own
sentence.

On the other hand, Alt-8 is rated poorly by User B; it violates B's
preferences for linear order (remember that Rules 21 and 22 specify
that B prefers the claim first, followed by cuisine information).  Also,
B's model has rules that radically decrease the
ranking of examples using the {\sc period-infer} operation (Rules 2, 4, 7 and 9).

Thus, Alt-6 and Alt-8 show that users A and B prefer different
combination operators, and different ordering of content, e.g. B likes
the claim first and A likes recommendations with food quality first
followed by cuisine. As mentioned above, previous work on the
generation of evaluative arguments states that the claim may appear
first or last \fullcite{careninimoore00b}. The relevant guideline for
producing effective evaluative arguments states that ``placing the
main claim first helps users follow the line of reasoning, but
delaying the claim until the end of the argument can also be effective
if the user is likely to disagree with the claim.'' The template-based
generator for {\sc MATCH} always placed the claim first, but this analysis
suggests that this may not be effective for user A.

{\bf Other similarities and differences:} There are also individual
differences in preferences for particular operations, and for specific
content operation interactions. For example, User A's model demotes
examples where the {\sc with-ns} operation has been applied (Rules 3,
6 and 8 in Figure~\ref{amanda-rule-fig}), while User B generally likes
examples where {\sc with-ns} has been used (Rule 18 in
Figure~\ref{lyn-rule-fig}). However, neither A nor B like {\sc
with-ns} when used to combine other content with neighborhood
information. In User A's model the $\alpha$ value is -1.26, while in
User B's model the value is -0.50 (see Rule 1 in
Figure~\ref{amanda-rule-fig} and Rule 3 in Figure~\ref{lyn-rule-fig}.)
These rules capture a specific interaction in the sp-tree between
domain-specific content and the {\sc with-ns-infer} combination operation.  Utterances
instantiating these rules place information in an adjunctival
with-clause following the clause realizing the restaurant's
neighborhood. There is no constraint on the type of information in the
with-clause.  In utterance (\ex{1}) below, the with-clause realizes
the restaurant's food quality, whereas in (\ex{2}) it contains
information about the restaurant's service.


\enumsentence{\label{nbhd-withns1} Mont Blanc has very good service, its price is 34 dollars, and it
 is located in Midtown West, with good food quality. It has the best overall quality among the selected restaurants.}

\enumsentence{\label{nbhd-withns2} Mont Blanc is located in Midtown West, with very good service, its price is 34 dollars, and it has good food quality. It has the best overall quality among the selected restaurants. }

Moreover, both users like {\sc with-ns} when it combines cuisine and
food-quality information as in example (\ex{1}) (Rule 23 in
Figure~\ref{amanda-rule-fig} and Rule 24 in
Figure~\ref{lyn-rule-fig}).



\enumsentence{\label{inter3} Komodo has the best overall quality among the selected restaurants since it is a Japanese, Latin American restaurant, with very good food quality, it has very good service, and its price is 29 dollars.}

But User B radically reduces the rating of the cuisine, food-quality
combination when it is combined with further information using the
{\sc relative-clause-infer} operation, as in example (\ex{2}) (Rules 5
and 6 in Figure~\ref{lyn-rule-fig}).

\enumsentence{\label{inter4} Bond Street has very good decor. This Japanese, Sushi restaurant, with excellent food quality, has good service. It has the best overall quality among the selected restaurants.}

Example (\ref{inter4}) is an interesting contrast with example
(\ref{inter3}).  Example (\ref{inter4}) instantiates Rule 24 in
Figure~\ref{lyn-rule-fig}, but it also instantiates a number of
negatively valued features. As discussed above, User B prefers
examples where the claim is expressed first (Rule 21 in
Figure~\ref{lyn-rule-fig}), and User B's model explicitly reduces the
rating of examples where information about decor is expressed first
(Rule 10 in Figure~\ref{lyn-rule-fig}).

In general, User A likes the {\sc merge-infer} operation (Rules 19, 21
and 22), especially when applied with {\sc assert-reco-food-quality}
(Rule 18), and {\sc assert-reco-decor} (Rule 20). User A strongly
prefers to hear about food quality first (Rule 26 in
Figure~\ref{amanda-rule-fig}), followed by cuisine information (Rule
24). In contrast, User B has rules that reduce the rating of examples
with price or decor first (Rules 8 and 10 in
Figure~\ref{lyn-rule-fig}). User B also has no preferences for {\sc
merge-infer} but likes the {\sc cw-conjunction} operation (Rule 20 in
Figure~\ref{lyn-rule-fig}).  Finally, User B dislikes the {\sc
relative-clause-infer} operation in general (Rule 1), and its
combination with the {\sc with-ns} operation (Rule 12) or the {\sc
period-infer} operation (Rule 11).

In addition to other evidence discussed above as to individual
differences in language generation, we believe that the fact that
these model differences are {\it interpretable} shows that the
differences in user perception of the quality of system utterances are
true individual differences, and not random noise.


\subsection{Average Model Differences}
\label{average-sec}

\begin{figure}[htb]
\begin{center}
\begin{small}\begin{tabular}{|c p{5.1in} c|}
\hline
N & Condition & $\alpha_s$ \\
\hline
1 & {\sc s-anc-attr-with*locate} $\geq -\infty$ & -0.87 \\ \hline
2 & {\sc s-trav-have1*i-restaurant*cuisine-type*ii-quality*attr-good*attr-food} $\geq -\infty$ & -0.81 \\ \hline
3 & {\sc s-trav-propernoun-restaurant} $\geq 2.5$ & -0.81 \\ \hline
4 & {\sc r-anc-cw-conjunction-infer*cw-conjunction-infer*period-justify} $\geq -\infty$ & -0.77 \\ \hline
5 & {\sc r-sis-assert-reco-relative-clause-infer} $\geq -\infty$ & -0.74 \\ \hline
6 & {\sc r-anc-assert-reco-decor*with-ns-infer*period-infer*period-justify} $\geq -\infty$ & -0.62 \\ \hline
7 & {\sc r-anc-assert-reco-cuisine*with-ns-infer*relative-clause-infer} $\geq -\infty$ & -0.62 \\ \hline
8 & {\sc period-justify-avg-leaves-under} $\geq 5.5$ & -0.60 \\ \hline

9 & {\sc cw-conjunction-infer-avg-leaves-under} $\geq 3.1$ & -0.54 \\ \hline
10 & {\sc r-anc-assert-reco-nbhd*with-ns-infer} $\geq -\infty$ & -0.45 \\ \hline
11 & {\sc r-sis-cw-conjunction-infer-relative-clause-infer} $\geq -\infty$ & -0.40 \\ \hline
 12& {\sc period-infer-avg-leaves-under} $\geq 3.4$ & 0.14 \\ \hline
 13& {\sc r-anc-assert-reco-food-quality*merge-infer} $\geq -\infty$ & 0.15 \\ \hline
 14& {\sc s-trav-propernoun-restaurant} $\geq 5.5$ & 0.19 \\ \hline
 15& {\sc r-anc-assert-reco-decor*merge-infer} $\geq -\infty$ & 0.19 \\ \hline

 16& {\sc s-anc-attr-with*i-restaurant*have1} $\geq -\infty$ & 0.22 \\ \hline
 17& {\sc r-anc-assert-reco-decor*with-ns-infer} $\geq -\infty$ & 0.26 \\ \hline
 18& {\sc leaf-assert-reco-best*assert-reco-cuisine*assert-reco-service} $\geq -\infty$ & 0.26 \\ \hline
 19& {\sc s-trav-propernoun-restaurant} $\geq 3.5$ & 0.29 \\ \hline
 20& {\sc r-anc-assert-reco-cuisine*with-ns-infer*period-infer*period-justify} $\geq -\infty$ & 0.29 \\ \hline
 21& {\sc leaf-assert-reco-food-quality} $\geq -\infty$ & 0.32 \\ \hline
 22& {\sc period-infer-avg-leaves-under} $\geq 3.2$ & 0.36 \\ \hline
 23& {\sc r-sis-merge-infer-assert-reco} $\geq -\infty$ & 0.42 \\ \hline
 24& {\sc period-justify-avg-leaves-under} $\geq 6.5$ & 0.48 \\ \hline
25& {\sc s-anc-attr-with*have1} $\geq -\infty$ & 0.49 \\ \hline
 26& {\sc leaf-assert-reco-best*assert-reco-cuisine} $\geq -\infty$ & 0.50 \\ \hline
 27& {\sc leaf-assert-reco-best} $\geq -\infty$ & 0.50 \\ \hline
 28& {\sc merge-infer-max-leaves-under} $\geq -\infty$ & 0.51 \\ \hline
 29& {\sc leaf-assert-reco-food-quality*assert-reco-cuisine} $\geq -\infty$ & 0.77 \\ \hline
 30& {\sc merge-infer-max-leaves-under} $\geq 2.5$ & 0.96 \\ \hline
 31& {\sc s-trav-have1*propernoun-restaurant*ii-quality*attr-among1} $\geq -\infty$ & 0.97 \\ \hline
\end{tabular}\end{small}
\end{center}
\caption{A subset of the rules with the largest $\alpha$ magnitudes that were learned for ranking recommendations given AVG feedback. \label{rec-avg-rules}}
\end{figure}

Table~\ref{rec-avg-rules} shows a subset of rules that have the
largest $\alpha$ magnitudes for an example AVG model using the same
100 feature bootstrapping process described above.
Section~\ref{indiv-sec} presented results that the average model
performs statistically worse for recommendations than either of the
individual models. This may be due to the fact that the average model
is essentially trying to learn from contradictory feedback from the
two users.  To see whether an examination of the models provides
support for this hypothesis, we first examine how the learned model
ranks Alt-6 And Alt-8 as shown in Figure~\ref{recommend-alt-spr-fig}
in the column SPR$_{AVG}$.  The average feedback for Alt-6 is 2.5
while the average feedback for Alt-8 is 3, but the trained {\spr}
ranks Alt-8 second highest and Alt-6 fifth out of 10.



The mid-value ranking of Alt-6 arises from a number of interacting
rules, some of which are similar to User B's and some of which are
similar to User A's. Alt-6 instantiates Rules 26 and 27 in
Figure~\ref{rec-avg-rules} which increase the ranking of sentence
plans in which the claim, i.e. {\sc assert-reco-best} is realized
first, and sentence plans where the claim is immediately followed by
information about the type of cuisine ({\sc
assert-reco-cuisine}). These rules are identical to B's Rules 21 and
22 in Figure~\ref{lyn-rule-fig}. Rule 18 additionally increases the
ranking of sentence plans where cuisine information is followed by
service information, which applies to Alt-6 to further increase its
ranking.  However Rule 3 lowers the ranking of Alt-6, since it
combines more than 3 different assertions into a single DSyntS tree.

Alt-8 is highly ranked by SPR$_{AVG}$, largely as a result of several
rules that increase its ranking. Rule 31 specifies an increase in
ranking for sentence plans that have the claim in its own sentence,
which is true of Alt-8 but not of Alt-6. This rule also appears as
Rule 27 in A's model in Figure~\ref{amanda-rule-fig}. Alt-8 also
instantiates Rules 21 and 29 which which are identical to user A's
ordering preferences (Rules 24 and 26 in Figure~\ref{amanda-rule-fig})
These rules describe sp-trees with {\sc assert-reco-food-quality} on
the left frontier, and trees where it is followed by {\sc
assert-reco-cuisine}. (See Alt-8's sp-tree in
Figure~\ref{alt8-sptree}.) Rule 3 also applies to Alt-8, reducing its
ranking due to the number of content items it realizes.

{\bf Other similarities and differences}: There are many rules in the
average model that are similar to either A or B's models or both, and
the average model retains a number of preferences seen in the
individual models. For example, Rules 1 and 10 both reduce the
ranking of any sentence plan where neighborhood information is
combined with subsequent information using the {\sc with-ns}
combination operator. Rule 1 expresses this in terms of the lexical
items in the d-tree, whereas Rule 10 expresses it in terms of
semantic features derived from the sp-tree. Examples
\ref{nbhd-withns1} and \ref{nbhd-withns2} in Section~\ref{indiv-sec} illustrate this interaction.

Some of the rules are more similar to User A. For example, Rules 4 and
9 (like A's Rules 2 and 5 in Figure~\ref{amanda-rule-fig}) reduce the
rating of sentence plans that use the operation {\sc cw-conj-infer}.
In addition, Rules 22, 23, 24, and 28 expresses preferences for
merging information, which are very similar to A's Rules 19, 21 and
22.  Rule 15 expresses a preference for information about the
atmosphere ({\sc assert-reco-decor}) to be combined using the {\sc
merge} operation, as specified in A's Rule 20. Rule 20 in
Figure~\ref{rec-avg-rules} is also similar to A's Rule 16
with {\sc assert-reco-cuisine} combined with subsequent information
with the {\sc with-ns} operation.

Other rules are more similar to B's model. For example, Rule 5 reduces
the ranking of sentence plans using the {\sc relative clause}
operation, which was also specified in User B's Rule 1, and Rules 16
and 25 indicate a general preference for use of the with-ns operation,
which was a strong preference in User B's model (see B's Rule 18 in
Figure~\ref{lyn-rule-fig}).

Note that in some cases, the learned model tries to account for both
A's and B's preferences, even when these contradict one another. For
example, Rule 27 specifies a preference for the claim to come first,
as in B's Rule 21, whereas Rule 26 is the same as A's 24, specifying a
preference for food quality and cuisine information to be expressed
first.  Thus the model does
suggest that a reduction in performance may arise from trying to
account for the contradictory preferences of users A and B.

\section{Conclusions}
\label{discuss-sec}

This article describes {\sc SPaRKy}, a two-stage sentence planner that
generates many alternative realizations of input content plans and
then ranks them using a statistical model trained on human feedback.
We demonstrate that the training technique developed for SPoT
\fullcite{WRR02}, generalizes easily to new domains, and that it can be
extended to handle the rhetorical structures required for more complex
types of information presentation.

One of the most novel contributions of this paper is to show that
trainable generation can be used to train sentence planners tailored
to individual users' preferences. Previous work modeling individuals
has mainly applied to content planning.  While studies of human-human
dialogue suggest that modeling other types of individual differences
could be valuable for spoken language generation, in the past,
linguistic variation among individuals was considered a problem for
generation \fullcite{McKeownetal94,Reiter02,Reiteretal03}.  Here, we show
that users have different perceptions of the quality of alternative
realizations of a content plan, and that individualized models perform
better than those trained for groups of users.  Our qualitative
analysis indicates that trainable sentence generation is sensitive to
variations in domain application, presentation type, and individual
human preferences about the arrangement of particular content types.
These are the first results showing that individual preferences apply
to sentence planning.

We also compared {\sc SPaRKy} to the template-based generator described in
Section~\ref{template-sec}: this generator is highly tuned to this
domain and was previously shown to produce high quality outputs in a
user evaluation \fullcite{Stentetal04}. When {\sc SPaRKy} is trained for a group
of users, then template-based generation is better for {\sc recommend}
and {\sc compare-3}, but in most cases the performance of the
individualized {\spr}s are statistically indistinguishable from
{\sc MATCH}'s template-based generator: the exceptions are that, for {\sc
compare-2}, User B prefers {\sc SPaRKy}, while for {\sc compare-3} User A
prefers the template-based generator.  In all cases, the Human scores
(outputs produced by the {\spg} but selected by a human) are as good
or better than the template-based generator, even for complex
information presentations such as extended comparisons.

These results show that there is a gap between the performance of the
trained {\spr} and human performance.  This suggests that it might be
possible to improve the {\spr} with different feature sets or a
different ranking algorithm. We leave a comparison with other ranking
algorithms to future work. Here, we report results for many different
feature sets (n-gram, concept and tree) and investigate their effect
on performance. Table~\ref{ranking-loss-feats} shows that a
combination of the three feature sets performs significantly better
for {\sc recommend} and {\sc compare-3} than the tree features from
our earlier work \fullcite{WRR02,Stentetal04,MairesseWalker05}.
Interestingly, in some cases, simple features like n-grams perform as
well as features representing linguistic structure such as the tree
features.  This might be because particular lexical items, e.g. {\it
with}, are often uniquely associated with a combination operator,
e.g. the {\sc with-ns} operator, which was shown to have impact on
user perceptions of utterance quality
(Section~\ref{qual-results-sec}). More work is needed to determine
whether these performance similarities are simply due to the fact that
the variation of form generated by {\sc SPaRKy}'s {\spg} is limited.  Other
work has also examined tradeoffs between n-gram features and
linguistically complex features in terms of tradeoffs between time and
accuracy \fullcite{Panteletal04}. Although {\sc SPaRKy} is trained offline, the
time to compute features and rank {\spg} outputs remains an issue when
using {\sc SPaRKy} in a real-time spoken dialogue system.

A potential limitation of our approach is the time and effort required
to elicit user feedback for training the system, as described in
Section~\ref{spr-sec}.  In Section~\ref{individual-results-sec} we
showed that RankLoss error rates of around 0.20 could be acquired with
a much smaller training set, i.e. with a training set of around 120
examples.  However typical users would probably not want to provide
ratings of 120 examples. Future work should explore alternative
training regimes perhaps by utilizing ratings from several users.  For
example, we could identify examples that most distinguish our existing
users, and just present these examples to new users.  Also, instead of
users rating information presentations before using {\sc MATCH}, perhaps a
method for users to rate information presentations while using {\sc MATCH}
could be developed, i.e. in the course of a dialogue with {\sc MATCH} when a
recommendation or comparison is presented to the user, the system
could display on the screen a rating form for that presentation.
Another approach would be to train from a different type of user
feedback collected automatically by monitoring the user's behavior,
e.g. measures of cognitive load such as reading time.

Another limitation is that {\sc SPaRKy}'s dictionary is handcrafted,
i.e. the associations between simple assertions and their syntactic
realizations (d-trees) are specified by hand, like all
generators. Recent work has begun to address this limitation by
investigating techniques for learning a generation dictionary
automatically from different types of corpora, such as user reviews
\fullcite{barzilay:emnlp02,higashinakawalkerprasad07,snyder:ijcai07}.

A final limitation is that we only use two individuals to provide a
proof-of-concept argument for the value of user-tailored trainable
sentence planning. We have argued throughout this paper that the
individual differences we document are more general, are not
particular to users A and B, and are not the result of random noise in
user feedback.  Nevertheless, we hope that future work will test these
results against a larger population of individuals in order to provide
further support for these arguments and in order to characterize the full
range of individual differences in preferences for language variation in
dialogue interaction.

\begin{acks}

This work was partially funded by a DARPA Communicator Contract MDA972-99-3-0003,
by a Royal Society Wolfson Research Merit Award to M. Walker, and by a Vice
Chancellor's studentship to F. Mairesse.
\end{acks}




\bibliographystyle{theapa}

\bibliography{sparky-cl-22}

\end{document}